%% file: main.tex
\documentclass[utf8]{frontiersFPHY}
\pdfoutput=1
\usepackage{hyperref}       
\hypersetup{
    colorlinks=true,
    linkcolor=cyan,
    filecolor=magenta,      
    urlcolor=blue,
}
\usepackage{url}            
\usepackage{booktabs}       
\usepackage{amsmath}       
\usepackage{nicefrac} 
\usepackage{microtype} 
\usepackage{lipsum}
\usepackage{graphicx}
\usepackage{amsmath}
\usepackage{multirow}
\usepackage{xspace}
\usepackage[onehalfspacing]{setspace}
\usepackage{glossaries-prefix}
\usepackage[capitalise,noabbrev]{cleveref}
\usepackage{todonotes}
\usepackage{float}

\input{acronyms}

\def\keyFont{\fontsize{8}{11}\helveticabold }
\def\firstAuthorLast{Hawks {et~al.}} 
\def\Authors{Benjamin Hawks\,$^{1}$, Javier Duarte\,$^{2}$, Nicholas J. Fraser\,$^{3}$, Alessandro Pappalardo\,$^{3}$, Nhan Tran\,$^{1,4}$, Yaman Umuroglu\,$^{3}$
  }
\def\Address{$^{1}$Fermi National Accelerator Laboratory, Batavia, IL, United States $^{2}$University of California San Diego, La Jolla, CA, United States, $^{3}$Xilinx Research, Dublin, Ireland, $^{4}$Northwestern University, Evanston, IL, United States}

\def\corrAuthor{Nhan Tran}
\def\corrEmail{ntran@fnal.gov}

\newcommand{\unit}[1]{\,\text{#1}\xspace}
\newcommand{\hlsfml}{\texttt{hls4ml}\xspace}
\newcommand{\qkeras}{\textsc{QKeras}\xspace}
\newcommand{\brevitas}{\textsc{Brevitas}\xspace}
\newcommand{\norm}[1]{\|#1\|}
\renewcommand{\vec}[1]{\boldsymbol{#1}}

\newcommand{\eb}{\ensuremath{\epsilon_b^{\epsilon_s=0.5}}\xspace}
\DeclareMathOperator*{\clamp}{clamp}
\DeclareMathOperator*{\round}{round}

\begin{document}

\title[Quantization-Aware Pruning]{Ps and Qs: Quantization-Aware Pruning for Efficient Low Latency Neural Network Inference}

\firstpage{1}

\author[\firstAuthorLast ]{\Authors} 
\address{} 
\correspondance{} 

\extraAuth{}

\maketitle

\begin{abstract}
\section{}
Efficient machine learning implementations optimized for inference in hardware have wide-ranging benefits, depending on the application, from lower inference latency to higher data throughput and reduced energy consumption.  
Two popular techniques for reducing computation in neural networks are pruning, removing insignificant synapses, and quantization, reducing the precision of the calculations.  
In this work, we explore the interplay between pruning and quantization during the training of neural networks for ultra low latency applications targeting high energy physics use cases.  
Techniques developed for this study have potential applications across many other domains. 
We study various configurations of pruning during quantization-aware training, which we term \emph{quantization-aware pruning}, and the effect of techniques like regularization, batch normalization, and different pruning schemes on performance, computational complexity, and information content metrics.
We find that quantization-aware pruning yields more computationally efficient models than either pruning or quantization alone for our task. 
Further, quantization-aware pruning typically performs similar to or better in terms of computational efficiency compared to other neural architecture search techniques like Bayesian optimization.  
Surprisingly, while networks with different training configurations can have similar performance for the benchmark application, the information content in the network can vary significantly, affecting its generalizability. 
\tiny
 \keyFont{ \section{Keywords:} pruning, quantization, neural networks, generalizability, regularization, batch normalization} 
\end{abstract}


\section{Introduction}
\input{intro}

\section{Benchmark task}
\input{task}

\section{Quantization-aware pruning}
\input{setup}

\section{Results}
\input{results}

\section{Summary and outlook}
\input{outlook}

\section*{Data Availability Statement}
Publicly available datasets were analyzed in this study. This data can be found here: \href{https://zenodo.org/record/3602254}{https://zenodo.org/record/3602254}.

\section*{Author Contributions}
BH performed all of the training and testing with input, advice, and documentation by JD, NF, AP, NT, and YU.

\section*{Funding}
BH and NT are supported by Fermi Research Alliance, LLC under Contract No. DE-AC02-07CH11359 with the U.S. Department of Energy (DOE), Office of Science,
Office of High Energy Physics and the DOE Early Career Research program under Award No. DE-0000247070. 
JD is supported by the DOE, Office of Science, Office of High Energy Physics Early Career Research program under Award No. DE-SC0021187.

This work was performed using the Pacific Research Platform Nautilus HyperCluster supported by NSF awards CNS-1730158, ACI-1540112, ACI-1541349, OAC-1826967, the University of California Office of the President, and the University of California San Diego's California Institute for Telecommunications and Information Technology/Qualcomm Institute. 
Thanks to CENIC for the 100\,Gpbs networks.

\section*{Acknowledgments}
We acknowledge the Fast Machine Learning collective as an open community of multi-domain experts and collaborators. 
This community was important for the development of this project.  
Thanks especially to Duc Hoang for enabling evaluations of post-training quantized \textsc{PyTorch} models using \texttt{hls4ml}.
We would also like to thank Nick Schaub and Nate Hotaling from NCATS/Axel Informatics for their insight on aIQ.

\bibliographystyle{frontiersSCNS} 
\bibliography{references}

\noindent\textbf{Conflict of Interest}: Authors NF, AP, and YU were employed by the company Xilinx Research. 
The remaining authors declare that the research was conducted in the absence of any commercial or financial relationships that could be construed as a potential conflict of interest.

\end{document}

%% file: acronyms.tex
\glsdisablehyper

\newacronym[prefix={an\space},prefixfirst={a\space}]{FPGA}{FPGA}{field-programmable gate array}

\newacronym{NN}{NN}{neural network}
\newacronym{ML}{ML}{machine learning}
\newacronym{QAT}{QAT}{quantization-aware training}
\newacronym{QAP}{QAP}{quantization-aware pruning}
\newacronym{PTQ}{PTQ}{post-training quantization}
\newacronym{FT}{FT}{fine-tuning}
\newacronym{LT}{LT}{lottery ticket}
\newacronym{BN}{BN}{batch normalization}
\newacronym{BO}{BO}{Bayesian optimization}
\newacronym{ROC}{ROC}{receiver operating characteristic}
\newacronym{AUC}{AUC}{area under the curve}

\newacronym{aiQ}{aiQ}{artificial intelligence quotient}
\newacronym{BOP}{BOP}{bit operation}
\newacronym{FLOP}{FLOP}{floating-point operation}

\newacronym{LHC}{LHC}{CERN Large Hadron Collider}

%% file: intro.tex
Efficient implementations of \gls{ML} algorithms provide a number of advantages for data processing both on edge devices and at massive data centers.  
These include reducing the latency of \gls{NN} inference, increasing the throughput, and reducing power consumption or other hardware resources like memory.  
During the ML algorithm design stage, the computational burden of \gls{NN} inference can be reduced by eliminating nonessential calculations through a modified training procedure. 
In this paper, we study efficient \gls{NN} design for an ultra-low latency, resource-constrained particle physics application.  
The classification task is to identify radiation patterns that arise from different elementary particles at sub-microsecond latency.  
While our application domain emphasizes low latency, the generic techniques we develop are broadly applicable.

Two popular techniques for efficient ML algorithm design are \emph{quantization} and \emph{pruning}.  
Quantization is the reduction of the bit precision at which calculations are performed in a \gls{NN} to reduce the memory and computational complexity.  
Often, quantization employs fixed-point or integer calculations, as opposed to floating-point ones, to further reduce computations at no loss in performance.  
Pruning is the removal of unimportant weights, quantified in some way, from the \gls{NN}.  
In the most general approach, computations are removed, or pruned, one-by-one from the network, often using their magnitude as a proxy for their importance. 
This is referred to as magnitude-based unstructured pruning, and in this study, we generically refer to it as pruning.  
Recently, \gls{QAT}, accounting for the bit precision at training time, has been demonstrated in a number of studies to be very powerful in efficient ML algorithm design.  
In this paper, we explore the potential of combining pruning with \gls{QAT} at any possible precision.  
As one of the first studies examining this relationship, we term the combination of approaches \emph{\gls{QAP}}.  
The goal is to understand the extent to which pruning and quantization approaches are complementary and can be optimally combined to create even more efficiently designed \glspl{NN}.

Furthermore, as detailed in Sec.~\ref{sec:related}, there are multiple approaches to efficient \gls{NN} optimization and thus also to \gls{QAP}.  
While different approaches may achieve efficient network implementations with similar classification performance, these trained \glspl{NN} may differ in their information content and computational complexity, as quantified through a variety of metrics.
Thus, some approaches may better achieve other desirable characteristics beyond classification performance such as algorithm robustness or generalizability.  

This paper is structured as follows.
\cref{sec:related} briefly recapitulates related work.
\cref{sec:task} describes the low latency benchmark task in this work related to jet classification at the \gls{LHC}.  
\cref{sec:qap} introduces our approach to \gls{QAP} and the various configurations we explore in this work.  
To study the joint effects of pruning and quantization, we introduce the metrics we use in \cref{sec:metrics}.
The main results are reported in \cref{sec:results}.
Finally, a summary and outlook are given in \cref{sec:outlook}.\\

\subsection{Related work}
\label{sec:related}
While \glspl{NN} offer tremendous accuracy on a variety of tasks, they typically incur a high computational cost.
For tasks with stringent latency and throughput requirements, this necessitates a high degree of efficiency in the deployment of the \gls{NN}.
A variety of techniques have been proposed to explore the efficient processing of \glspl{NN}, including quantization, pruning, low-rank tensor decompositions, lossless compression and efficient layer design.
We refer the reader to \cite{sze2020efficient} for a survey of techniques for efficient processing of \glspl{NN}, and focus on related work around the key techniques covered in this paper.

\textbf{Pruning.} Early work~\citep{optimalbraindamage} in \gls{NN} pruning identified key benefits including better generalization, fewer training examples required, and improved speed of learning the benefits through removing insignificant weights based on second-derivative information.
Recently, additional compression work has been developed in light of mobile and other low-power applications, often using magnitude-based pruning~\citep{han2016deep}.
In \cite{lotteryticket}, the authors propose the \emph{\gls{LT} hypothesis}, which posits that sparse subnetworks exist at initialization which train faster and perform better than the original counterparts.
\cite{learningraterewinding} proposes learning rate rewinding in addition to weight rewinding to more efficiently find the winning lottery tickets.
\cite{supermask} extends these ideas further to learning ``supermasks'' that can be applied to an untrained, randomly initialized network to produce a model with performance far better than chance.
The current state of pruning is reviewed in \cite{stateofpruning}, which finds current metrics and benchmarks to be lacking.

\textbf{Quantization.} Reducing the precision of a static, trained network's operations, \emph{\gls{PTQ}}, has been explored extensively in the literature~\citep{Duarte:2018ite,nagel2019datafree,han2016deep,meller2019same,zhao2019improving,banner2019posttraining}.
\gls{QAT}~\citep{bertmoons,NIPS2015_5647,zhang2018lq, ternary-16,zhou2016dorefa,JMLR:v18:16-456,xnornet, micikevicius2017mixed,Zhuang_2018_CVPR, wang2018training,bnnpaper} has also been suggested with different frameworks like \textsc{QKeras}~\citep{Coelho:2020zfu,qkeras} and \brevitas~\citep{blott2018finnr,brevitas} developed specifically to explore quantized NN training. 
Hessian-aware quantization (HAWQ)~\citep{hawq,hawqv2} is another quantization approach that uses second derivative information to automatically select the relative bit precision of each layer.
The Bayesian bits approach attempts to unify structured pruning and quantization by identifying pruning as the 0-bit limit of quantization~\citep{bayesianbits}.
In \cite{hacene2018quantized}, a combination of a pruning technique and a quantization scheme that reduces the complexity and memory usage of convolutional layers, by replacing the convolutional operation by a low-cost multiplexer, is proposed. In partuclar, the authors propose an efficient hardware architecture implemented on \gls{FPGA} on-chip memory.
In \cite{chang2020mix}, the authors apply different quantization schemes (fixed-point and sum-power-of-two) to different rows of the weight matrix to achieve better utilization of heterogeneous \gls{FPGA} hardware resources. 

\textbf{Efficiency metrics.} Multiple metrics have been proposed to quantify neural network efficiency, often in the context of dedicated hardware implementations.
The \gls{aiQ} is proposed in \cite{aiq} as metric to measure the balance between performance and efficiency of \glspl{NN}.
\Glspl{BOP}~\citep{bops} is another metric that aims to generalize \glspl{FLOP} to heterogeneously quantized \glspl{NN}.
A hardware-aware complexity metric (HCM)~\citep{hcm} has also been proposed that aims to predict the impact of \gls{NN} architectural decisions on the final hardware resources.
Our work makes use of some of these metrics and further explores the connection and tradeoff between pruning and quantization.

%% file: task.tex
\label{sec:task}


The LHC is a proton-proton collider that collides bunches of protons at a rate of 40\unit{MHz}.
To reduce the data rate, an online filter, called the trigger system, is required to identify the most interesting collisions and save them for offline analysis.
A crucial task performed on \glspl{FPGA} in the Level-1 trigger system that can be greatly improved by \gls{ML}, both in terms of latency and accuracy, is the classification of particles coming from each proton-proton collision. 
The system constraints require algorithms that have a latency of $\mathcal{O}$($\mu$s) while minimizing the limited \gls{FPGA} resources available in the system.  

\begin{figure*}
    \centering
    \includegraphics[width=0.7\textwidth]{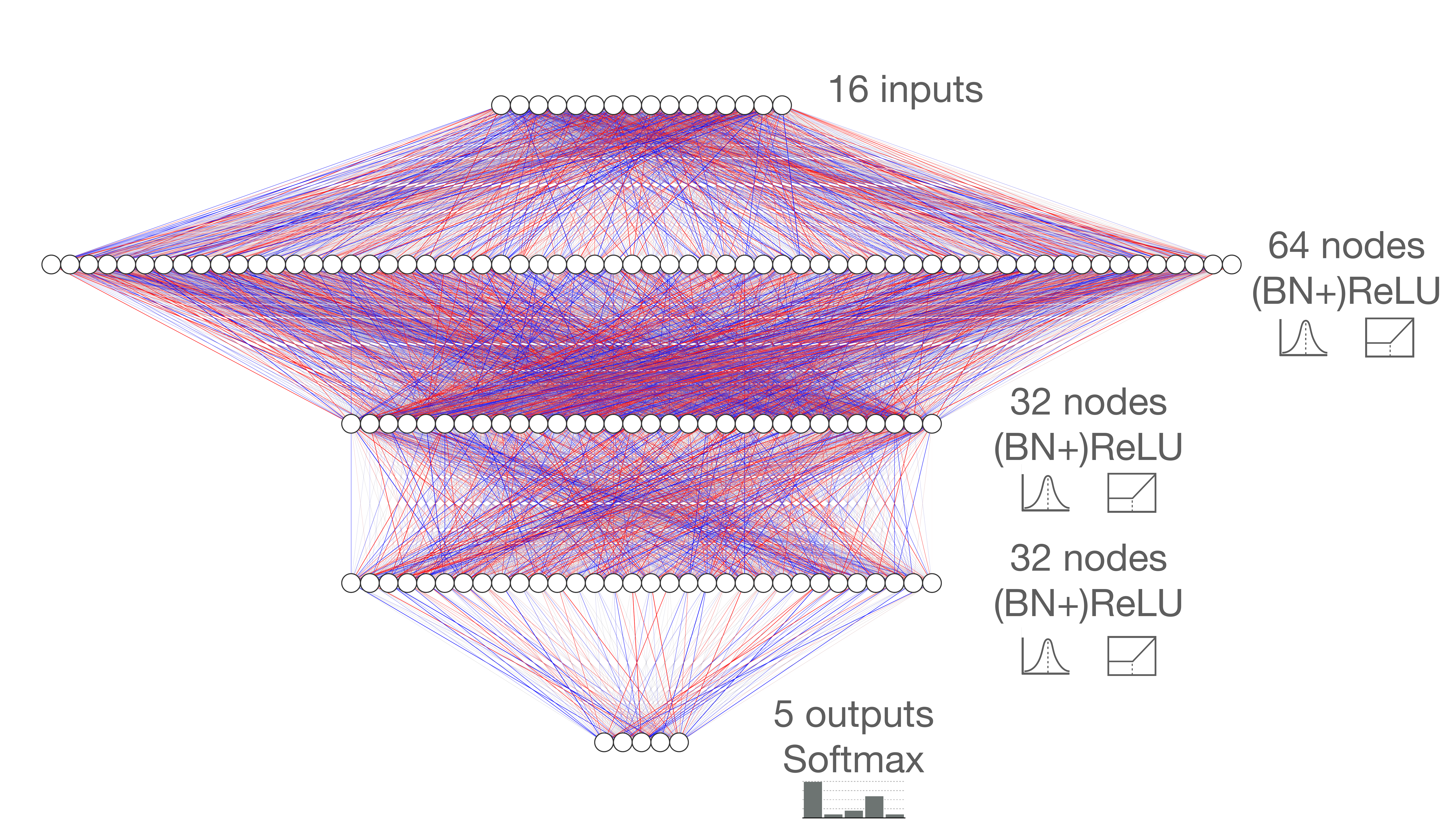}
    \caption{Baseline fully-connected neural network architecture, consisting of 16 inputs, five softmax-activated outputs, and three hidden layers. 
    The three hidden layers contain 64, 32, and 32 hidden nodes each with ReLU activation.
    A configuration with batch normalization (BN) layers before each ReLU activation function is also considered.
    The red and blue lines represent positive and negative weights, respectively, and the opacity represents the magnitude of each weight for this randomly initialized network.}
    \label{fig:architectures}
\end{figure*}

We consider a benchmark dataset for this task to demonstrate our proposed model efficiency optimization techniques.
In \cite{Coleman:2017fiq,Duarte:2018ite,Moreno:2019bmu}, a dataset~\citep{hls4mldata_100p} was presented for the classification of collimated showers of particles, or \emph{jets}, arising from the decay and hadronization of five different classes of particles: light flavor quarks (q), gluons (g), W and Z bosons, and top quarks (t). 
For each class, jets are pair-produced ($\mathrm{W}^+\mathrm{W}^-,\mathrm{Z}\mathrm{Z},\mathrm{q}\overline{\mathrm{q}},\mathrm{t}\overline{\mathrm{t}},\mathrm{g}\mathrm{g}$) in proton-proton collisions at a center-of-mass energy of 13\unit{TeV} from the same $\mathrm{q}\overline{\mathrm{q}}$ initial state.  
The jets are selected such that the unshowered parton or boson has a transverse momentum of 1\unit{TeV} within a narrow window of $\pm 1\% (10\unit{GeV})$ such that transverse momenta spectra is similar for all classes.  
Each jet is represented by 16 physics-motivated high-level features which are presented in Table~1 of \cite{Coleman:2017fiq}. 
The dataset contains 870,000 jets, balanced across all classes and split into 472,500 jets for training, 157,500 jets for validation, and  240,000 jets for testing.
Adopting the same baseline architecture as in \cite{Duarte:2018ite}, we consider a fully-connected \gls{NN} consisting of three hidden layers (64, 32, and 32 nodes, respectively) with rectified linear unit (ReLU)~\citep{relu1,relu2} activation functions, shown in \cref{fig:architectures}. 
The output layer has five nodes, yielding a probability for each of the five classes through a softmax activation function.
We refer to this network as the baseline floating-point model.


%% file: setup.tex


\label{sec:qap}
Applying quantization and pruning to a \gls{NN} can drastically improve its efficiency with little to no loss in performance. 
While applying these changes to a model post-training can be successful, to be maximally effective, we consider these effects at the time of \gls{NN} training.
Because computational complexity, as defined in Sec.~\ref{sec:metrics}, is quadratically dependent on precision while it is linearly dependent on pruning, the first step in our \gls{QAP} approach is to perform \gls{QAT}.  
This is followed by integrating pruning in the procedure.\\ 

\subsection{Quantization-aware training}

Quantized~\citep{JMLR:v18:16-456,DBLP:journals/corr/GongLYB14,wu2016quantized,37631,DBLP:journals/corr/GuptaAGN15,han2016deep} and even binarized~\cite{NIPS2015_5647,DBLP:journals/corr/GuptaAGN15,NIPS2016_6573,DBLP:journals/corr/RastegariORF16,DBLP:journals/corr/MerollaAAEM16,DiGuglielmo:2020eqx} \glspl{NN} have been studied as a way to compress \glspl{NN} by reducing the number of bits required to represent each weight and activation value.
As a common platform for \glspl{NN} acceleration, \glspl{FPGA} provide considerable freedom in the choice of data type and precision.  
Both choices should be considered carefully to prevent squandering \gls{FPGA} resources and incurring additional latency.
For example, in \qkeras and \hlsfml~\citep{Duarte:2018ite}, a tool for transpiling \glspl{NN} on \glspl{FPGA}, fixed-point arithmetic is used, which requires less resources and has a lower latency than floating-point arithmetic.  
For each parameter, input, and output, the number of bits used to represent the integer and fractional parts can be configured separately.
The precision can be reduced through \gls{PTQ}, where pre-trained model parameters are clipped or rounded to lower precision, without causing a loss in performance~\citep{DBLP:journals/corr/GuptaAGN15} by carefully choosing the bit precision.

Compared to \gls{PTQ}, a larger reduction in precision can be achieved through \gls{QAT}~\cite{bertmoons,NIPS2015_5647,ternary-16}, where the reduced precision of the weights and biases are accounted for directly in the training of the \gls{NN}.  
It has been found that \gls{QAT} models can be more efficient than \gls{PTQ} models while retaining the same performance~\citep{Coelho:2020zfu}. 
In these studies, the same type of quantization is applied everywhere. 
More recently~\citep{haq,hawq,hawqv2}, it has been suggested that per-layer heterogeneous quantization is the optimal way to achieve high accuracy at low resource cost. 
For the particle physics task with a fully-connected \gls{NN}, the accuracy of the reduced precision model is compared to the 32-bit floating-point implementation as the bit width is scanned.
In the \gls{PTQ} case~\citep{Duarte:2018ite}, the accuracy begins to drop below 14-bit fixed-point precision, while in the \gls{QAT} case implemented with \qkeras~\citep{Coelho:2020zfu} the accuracy is consistent down to 6 bits. 

In this work, we take a different approach to training quantized \glspl{NN} using \brevitas~\citep{brevitas}, a \textsc{PyTorch} library for \gls{QAT}. 
\brevitas provides building blocks at multiple levels of abstraction to compose and apply quantization primitives at training time. 
The goal of \brevitas is to model the data type restrictions imposed by a given target platform along the forward pass. Given a set of restriction, \brevitas provides several alternative learning strategies to fulfill them, which are exposed to the user as hyperparameters. 
Depending on the specifics of the topology and the overall training regimen, different learning strategies can be more or less successful at preserving the accuracy of the output \gls{NN}.
Currently, the available quantizers target variations of binary, ternary, and integer data types. 
Specifically, given a real valued input $x$, the integer quantizer $Q_\mathrm{int}(x)$ performs uniform affine quantization, defined as
\begin{equation}
Q_\mathrm{int}(x) = s \clamp_{y_\mathrm{min},y_\mathrm{max}} \left( \round \left( \frac{x}{s}\right)\right)
\label{eq:quant_function}
\end{equation}
where
\begin{equation}
\clamp_{y_\mathrm{min},y_\mathrm{max}}(y) = 
\begin{cases}
y_\mathrm{min} & y < y_\mathrm{min}\,, \\ 
y & y_\mathrm{min} \leq y \leq y_\mathrm{max}\,, \\ 
y_\mathrm{max} & y > y_\mathrm{max} \,,
\end{cases}
\label{eq:clamp_function}
\end{equation}
$\round(\cdot): \mathbb{R} \to \mathbb{Z}$ is a rounding function, $s \in \mathbb{R}$ is the \emph{scale factor}, and $y_\mathrm{min} \in \mathbb{Z}$ and $y_\mathrm{max} \in \mathbb{Z}$ are the minimum and maximum thresholds, respectively, which depend on the available word length (number of bits in a word).

In this work, we adopt round-to-nearest as the $\round$ function, and perform per-tensor quantization on both weights and activations, meaning that $s$ is constrained to be a scalar floating-point value.
As the ReLU activation function is used throughout, unsigned values are used for quantized activations.
Thus, for a word length of $n$, the clamp function, $\clamp_{A_\mathrm{min},A_\mathrm{max}}(\cdot)$, is used with $A_\mathrm{min} = 0$ and $A_\mathrm{max} = 2^n-1$.
Quantized weights are constrained to symmetric signed values so $\clamp_{w_\mathrm{min},w_\mathrm{max}}(\cdot)$ is used with $w_\mathrm{max} = 2^{n-1}-1$ and $w_\mathrm{min} = -w_\mathrm{max}$.

In terms of learning strategies, we apply the straight-through estimator (STE)~\citep{NIPS2015_5647} during the backward pass of the rounding function, which assumes that quantization acts as the identity function, as is typically done in \gls{QAT}. 
For the weights' scale, similar to \cite{jacob2018quantization}, $s_w$ is re-computed at each training step such that the maximum value in each weight tensor is represented exactly
\begin{equation}
s_w = \frac{\max_\mathrm{tensor}\left(|\vec{W}|\right)}{2^{n-1}-1}\,,
\label{eq:weight_scale_factor}
\end{equation}
where $\vec{W}$ is the weight tensor for a given layer and $\max_\mathrm{tensor}(\cdot)$ is the function that takes an input tensor and returns the maximum scalar value found within. 
For the activations, the scale factor $s_A$ is defined as:
\begin{equation}
s_{A} = \frac{s_{A,\mathrm{learned}}}{2^{n-1}}\,,
\label{eq:act_scale_factor}
\end{equation}
where $s_{A,\mathrm{learned}}$ is a parameter individual to each quantized activation layer, initialized to 6.0 (in line with the $\mathrm{ReLU}6(\cdot)$ activation function), and learned by backpropagation in logarithmic scale, as described in \cite{jain2019trained}.
In the following, we refer to this scheme as scaled-integer quantization.

\vspace{1cm}

\subsection{Integrating pruning}


Network compression is a common technique to reduce the size, energy consumption, and overtraining of deep \glspl{NN}~\cite{han2016deep}. 
Several approaches have been successfully deployed to compress networks~\citep{DBLP:journals/corr/abs-1710-09282,9043731,review2020}.
Here we focus specifically on \emph{parameter pruning}: the selective removal of weights based on a particular ranking~\citep{lotteryticket,learningraterewinding,stateofpruning,han2016deep,NIPS1989_250,2017arXiv171201312L}. 

Prior studies~\citep{Duarte:2018ite} have applied pruning in an iterative fashion: by first training a model then removing a fixed fraction of weights per layer then retraining the model, while masking the previously pruned weights.
This processed can be repeated, restoring the final weights from the previous iteration, several times until reaching the desired level of compression.
We refer to this method as \gls{FT} pruning.
While the above approach is effective, we describe here an alternative approach based on the \gls{LT} hypothesis~\citep{lotteryticket} where the remaining weights after each pruning step are initialized back to their original values (``weight rewinding'').
We refer to this method as \gls{LT} pruning.
We propose a new hybrid method for constructing efficient \glspl{NN}, \gls{QAP}, which combines a pruning procedure with training that accounts for quantized weights.  
As a first demonstration, we use \brevitas~\citep{brevitas} to perform \gls{QAT} and iteratively prune a fraction of the weights following the \gls{FT} pruning method. 
In this case, we \gls{FT} prune approximately 10\% of the original network weights (about 400 weights) each iteration, with a reduction in the number of weights to prune once a sparsity of 90\% is reached.
Weights with the smallest $L_1$ norms across the full model are removed each iteration. 

\begin{figure*}[tbh!]
    \centering
    \includegraphics[width=0.45\textwidth]{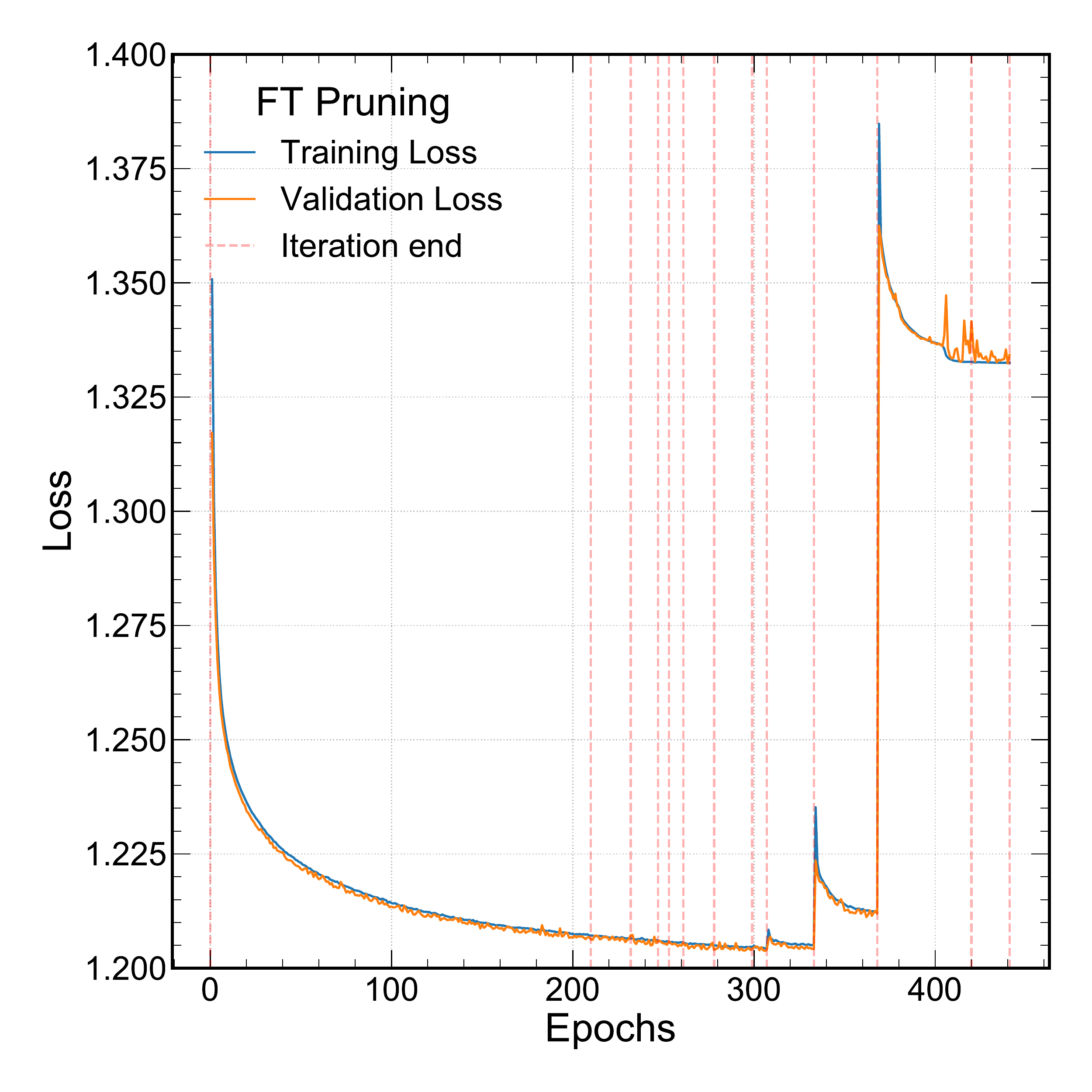}
    \includegraphics[width=0.45\textwidth]{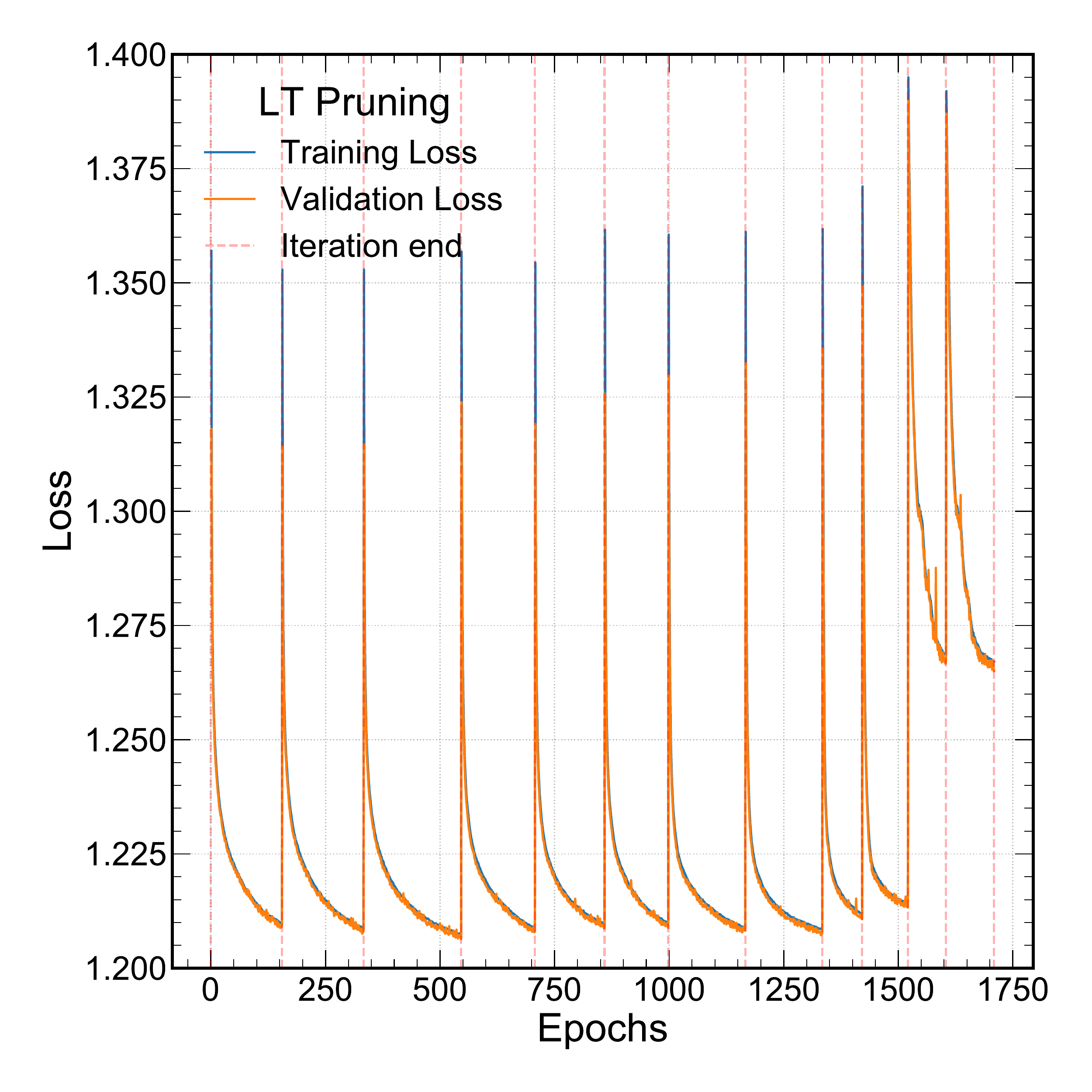}
    \caption{The loss function for the \gls{QAP} procedure for a 6-bit jet classification neural network. 
    \gls{FT} pruning is demonstrated on the left (A) and \gls{LT} pruning is shown on the right (B).}
    \label{fig:pruning_loss}
\end{figure*}

Our procedure for \gls{FT} and \gls{LT} pruning are demonstrated in \cref{fig:pruning_loss}, which shows the training and validation loss as a function of the epoch.  
To demonstrate the effect of \gls{QAP}, we start by training a network using \gls{QAT} for our jet substructure task constraining the precision of each layer to be 6 bits using \brevitas.  
This particular training includes \gls{BN} layers and $L_1$ regularization described in more detail in \cref{sec:opts}, although we also present results without these aspects. 

In \cref{fig:pruning_loss}A, the \gls{FT} pruning procedure iteratively prunes the 6-bit weights from the network.  
Each iteration is denoted by the dotted red lines after which roughly 10\% of the lowest magnitude weights are removed.  
At each iteration, we train for 250 epochs with an early stopping criteria of no improvement in the validation loss for 10 epochs.
The \gls{FT} pruning procedure continues to minimize or maintain the same loss over several pruning iterations until the network becomes so sparse that the performance degrades significantly around epoch 300.  
In \cref{fig:pruning_loss} (right), the \gls{LT} pruning procedure is shown.  
Our approach deviates from the canonical \gls{LT} pruning study~\citep{lotteryticket} in that we fully train each pruning iteration until the early stopping criteria is satisfied instead of partially optimizing the network. 
This is because we would like to explore the performance of the network at each stage of pruning to evaluate a number of metrics.  
However, the behavior is as expected---at each pruning iteration the loss goes back to its initial value.  
Similar to the \gls{FT} pruning case, when the \gls{LT} pruning neural network becomes very sparse, around epoch 1500, the performance begins to degrade.   
We note that because of the additional introspection at each iteration, our \gls{LT} pruning procedure requires many more epochs to train than the \gls{FT} pruning procedure.\\

\subsection{Neural network training configurations}
\label{sec:opts}
In this section, we describe \gls{BN} and $L_1$ regularization, which have the power to modify the efficiency of our \gls{QAP} models.
We also describe \gls{BO}, which we use to perform a standard neural architecture search for comparison to \gls{QAP}.\\

\subsubsection{Batch normalization and $L_1$ regularization}

\Gls{BN}~\citep{bn} was originally proposed to mitigate internal covariate shift, although others have suggested its true benefit is in improving the smoothness of the loss landscape~\citep{santurkar2019does}.
The \gls{BN} transformation $\vec{y}$ for an input $\vec{x}$ is
\begin{equation}
    \vec{y} = \vec{\gamma}\frac{\vec{x} - \vec{\mu}}{\sqrt{\vec{\sigma}^2 + \epsilon}} + \vec{\beta},
\end{equation}
given the running mean $\vec{\mu}$ and standard deviation $\vec{\sigma}$, the learnable scale $\vec{\gamma}$ and shift $\beta$ parameters, and $\epsilon$ a small number to increase stability.
Practically, the \gls{BN} layer shifts the output of dense layers to the range of values in which the activation function is nonlinear, enhancing the network's capability of modeling nonlinear responses, especially for low bit precision~\citep{bnnpaper,NIPS2015_5647}.
For this reason, it is commonly used in conjunction with extremely low bit precision.

We also train models with and without $L_1$ regularization~\citep{Duarte:2018ite,han2016deep,DBLP:journals/corr/HanPTD15}, in which the classification loss function $L_\mathrm{c}$ is augmented with an additional term, 
\begin{equation}
    L = L_\mathrm{c} + \lambda \norm{\vec{w}}_1\,,
\end{equation}
where $\vec{w}$ is a vector of all the weights of the model and $\lambda$ is a tunable hyperparameter.
This can be used to assist or accelerate the process of iterative pruning, as it constrains some weights to be small, producing already sparse models~\citep{Ng:2004:FSL:1015330.1015435}.
As the derivative of the penalty term is $\lambda$ whose value is independent of the weight, $L_1$ regularization can be thought of as a force that subtracts some constant from an ineffective weight each update until the weight reaches zero.\\

\subsubsection{Bayesian optimization}
\label{sec:bo}

\Gls{BO}~\citep{bo1,bo2,michael2010a} is a sequential strategy for optimizing expensive-to-evaluate functions.
In our case, we use it to optimize the hyperparameters of the neural network architecture.
\Gls{BO} allows us to tune hyperparameters in relatively few iterations by building a smooth model from an initial set of parameterizations (referred to as the ``surrogate model'') in order to predict the outcomes for as yet unexplored parameterizations. 
\Gls{BO} builds a smooth surrogate model using Gaussian processes (GPs) based on the observations available from previous rounds of experimentation. 
This surrogate model is used to make predictions at unobserved parameterizations and quantify the uncertainty around them. 
The predictions and the uncertainty estimates are combined to derive an acquisition function, which quantifies the value of observing a particular parameterization. 
We optimize the acquisition function to find the best configuration to observe, and then after observing the outcomes at that configuration a new surrogate model is fitted.
This process is repeated until convergence is achieved. 

We use the Ax and BoTorch libraries~\citep{ax-platform,balandat2019botorch,daulton2020differentiable} to implement the \gls{BO} based on the expected improvement (EI) acquisition function,
\begin{equation}
\mathrm{EI}(x) = \mathbb{E}\bigl[\min(f(x) - f^\ast), 0)\bigr]~,
\end{equation}
where $f^\ast = \min_i y_i$ is the current best observed outcome and our goal is to minimize $f$.
The total number of trials is set to 20 with a maximum number of parallel trials of 3 (after the initial exploration). 
Our target performance metric is the binary cross entropy loss as calculated on a ``validation'' subset of the jet substructure dataset. 
After the \Gls{BO} procedure is complete, and a ``best'' set of hyperparameters is found, each set of hyperparameters tested during the \Gls{BO} procedure is then fully trained for 250 epochs with an early stopping condition, and then metrics are calculated for each model on the ``test'' subset of the jet substructure dataset.


\section{Evaluation metrics}
\label{sec:metrics}
As we develop \gls{NN} models to address our benchmark application, we use various metrics to evaluate the \glspl{NN}' performance.  
Traditional metrics for performance include the classification accuracy, the \gls{ROC} curve of false positive rate versus true positive rate and the corresponding \gls{AUC}.
In physics applications, it is also important to evaluate the performance in the tails of distributions and we will introduce metrics to measure that as well.
The aim of quantization and pruning techniques is to reduce the energy cost of neural network implementations, and therefore, we need a metric to measure the computational complexity.  
For this, we introduce a modified version of \glspl{BOP}~\citep{bops}.
In addition, in this study we aim to understand how the network itself changes during training and optimization based on different neural network configurations.  
While the performance may be similar, we would like to understand if the information is organized in the neural network in the same way.  
Then we would like to understand if that has some effect on robustness of the model.  
To that end, we explore Shannon entropy metrics~\citep{shannon} and performance under class randomization.\\

\subsection{Classification performance}
\label{sec:perf}
\begin{figure}[tbh!]
    \centering
    \includegraphics[width=\columnwidth]{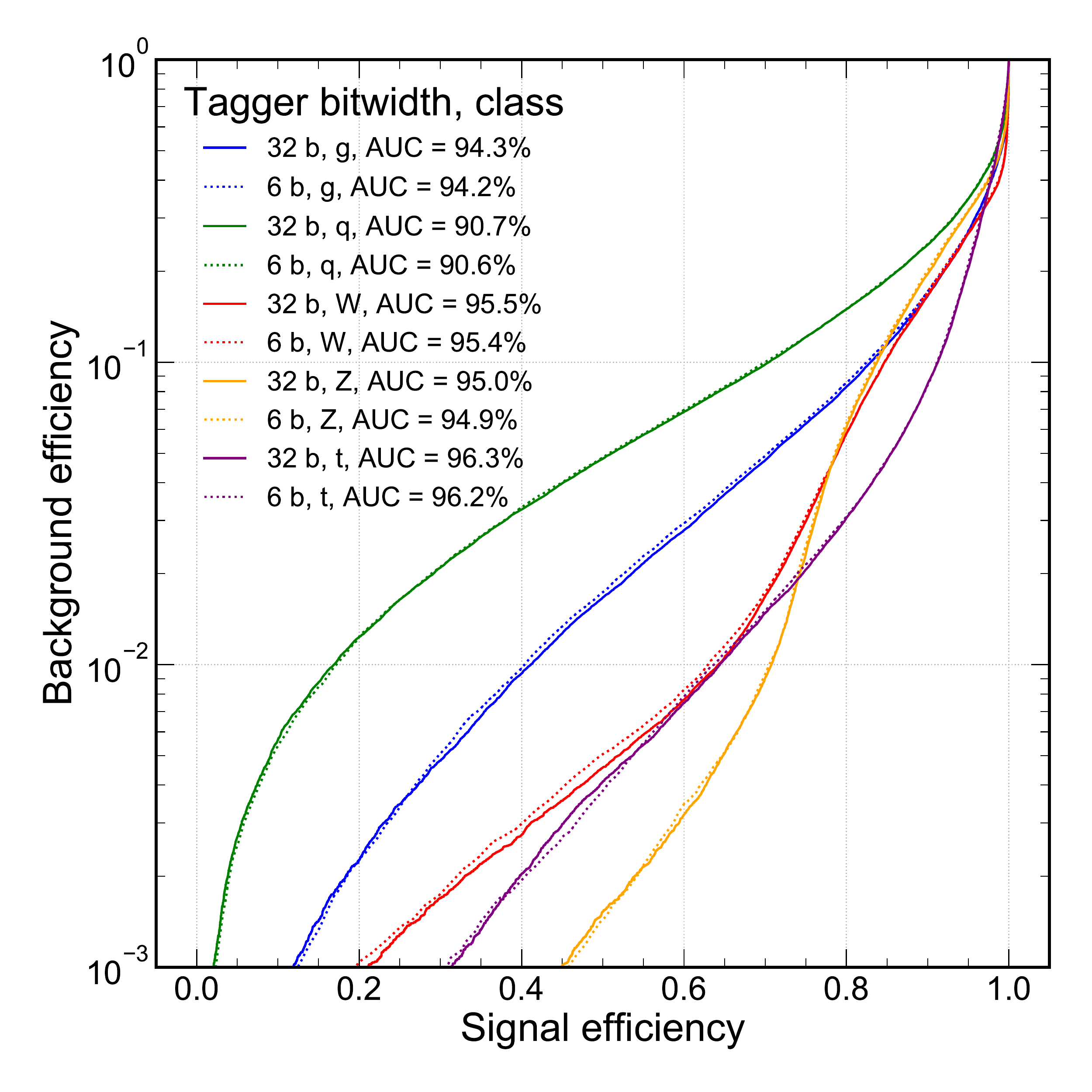}
    \caption{The ROC curve for each signal jet type class where the background are the other 4 classes.  Curves are presented for the unpruned 32-bit floating point classifier (solid lines) and 6-bit scaled integer models (dashed lines). All models are trained with batch normalization layers and $L_1$ Regularization.}
    \label{fig:fullROCs}
\end{figure}

For our jet substructure classification task, we consider the commonly-used accuracy metric to evaluate for the multi-class performance: average accuracy across the five jet classes.  
Beyond that, we also want to explore the full shape of the classifier performance in the \gls{ROC} curve.  
This is illustrated in \cref{fig:fullROCs} where the signal efficiency of each signal class is plotted against the misidentification probability for the other four classes, denoted as the background efficiency.  
The general features of \cref{fig:fullROCs} illustrate that gluon and quark jets are more difficult to distinguish than higher mass jet signals, W and Z boson, and the top quark.  
The Z boson is typically easier to distinguish than the W boson due to its greater mass.  
Meanwhile, the top quark is initially the easiest to distinguish at higher signal efficiency but at lower signal efficiencies loses some performance---primarily due to the top quark radiating more because the top quark has color charge.  
In particle physics applications, it is common to search for rare events so understanding tail performance of a classifier is also important.  
Therefore, as another performance metric, we define the background efficiency at a fixed signal efficiency of 50\%, $\eb$. 
We can report this metric $\eb$ for any signal type, considering all other classes as background processes.
From these \gls{ROC} curves, we see that $\eb$ can range from a few percent to the per-mille scale for the background samples.  
In Fig.~\ref{fig:fullROCs}, we show the \gls{ROC} curves for two \gls{NN} models: one trained with 32-bit floating-point precision and another one trained with \gls{QAT} at 6-bit scaled-integer precision.  
The networks are trained with $L_1$ regularization and \gls{BN} layers and do not include pruning. 

\subsection{Bit operations}
The goal of quantization and pruning is to increase the efficiency of the \gls{NN} implementation in hardware. 
To estimate the \gls{NN} computational complexity, we use the \glspl{BOP} metric~\citep{bops}. 
This metric is particularly relevant when comparing the performance of mixed precision arithmetic in hardware implementations on \glspl{FPGA} and ASICs.
We modify the \glspl{BOP} metric to include the effect of unstructured pruning.  
For a pruned fully-connected layer, we define it as
\begin{equation}
    \mathrm{BOPs} = mn \left[ (1-f_p)b_{\vec{a}}b_{\vec{w}} + b_{\vec{a}} + b_{\vec{w}} + \log_2(n) \right]
\end{equation}
where $n$ ($m$) is the number of inputs (outputs), $b_{\vec{w}}$ ($b_{\vec{a}}$) is the bit width of the weights (activations), and $f_p$ is the fraction of pruned layer weights.
The inclusion of the $f_p$ term accounts for the reduction in multiplication operations because of pruning.  
In the dominant term, due to multiplication operations ($b_{\vec{a}}b_{\vec{w}}$), \glspl{BOP} is quadratically dependent on the bit widths and linearly dependent on the pruning fraction. 
Therefore, reducing the precision is the first step in our \gls{QAP} procedure, as described above, followed by iterative pruning.\\

\subsection{Shannon entropy, neural efficiency, and generalizability}

Typically, the hardware-centric optimization of a \gls{NN} is a multi-objective, or Pareto, optimization of the algorithm performance (in terms of accuracy or \gls{AUC}) and the computational cost.  
Often, we can arrive at a range of Pareto optimal solutions through constrained minimization procedures.
However, we would like to further understand how the \textit{information} in different hardware-optimized \gls{NN} implementations are related.  
For example, do solutions with similar performance and computational cost contain the same information content?  
To explore that question, we use a metric called \emph{neural efficiency} $\eta_N$~\citep{aiq}.

Neural efficiency measures the utilization of state space, and it can be thought of as an entropic efficiency. 
If all possible states are recorded for data fed into the network, then the probability, $p_s$, of a state $s$ occurring can be used to calculate Shannon entropy $E_\ell$ of network layer $\ell$
\begin{equation}
E_\ell = -\sum_{s=1}^S p_s \log_2(p_s),
\end{equation}
where the sum runs over the total size of the state space $S$.
For a $b$-bit implementation of a network layer with $N_\ell$ neurons, this sum is typically intractable to compute, except for extremely low bit precision and small layer size, as the state space size is $S=2^{bN_\ell}$
Therefore, a simplification is made to treat the state of a single neuron as binary (whether the output value is greater than zero) so that $S=2^{N_\ell}$.
The maximum entropy of a layer corresponds to the case when all states occur with equal probability, and the entropy value is equal to the number of neurons $E_\ell = N_\ell$. 
The neural efficiency of a layer can then be defined as the entropy of the observed states relative to the maximum entropy: $\eta_\ell = E_\ell/N_\ell$.
Neuron layers with neural efficiency close to one (zero) are making maximal (minimal) usage of the available state space. 
Alternatively, high neural efficiency could also mean the layer contains too few neurons.

To compute the neural efficiency of a fully-connected NN $\eta_N$ we take the geometric mean of the neural efficiency of each layer $\eta_\ell$ in the network
\begin{equation}
\label{eq:neff}
\eta_N = \left(\prod_{\ell=1}^{L} \eta_\ell\right )^\frac{1}{L}    
\end{equation}
Although neural efficiency $\eta_N$ does not directly correlate with NN performance, in \cite{aiq}, it was found there was connection between NN generalizability and the neural efficiency.  
NNs with higher neural efficiency that maintain good accuracy performance were able to perform better when classes were partially randomized during training.  
The interpretation is that such networks were able to learn general features of the data rather than memorize images and therefore are less susceptible to performance degradation under class randomization.  
Therefore, in the results of our study, we also explore the effect of class randomization on our jet substructure task.

%% file: results.tex
\label{sec:results}


In the previous sections, we have introduced the benchmark task, the \gls{QAP} approach, and metrics by which we will evaluate the procedure.  In this section, we present the results of our experiments.  Our experiments are designed to address three conceptual topics:
\begin{itemize}
    \item In \cref{sec:res_perf}, we aim to study how certain training configuration choices can affect the performance (accuracy and \eb) of our \gls{QAP} procedure and how it compares to previous works.  
    In particular, we study the dependence of performance on the pruning procedure, the bit width, and whether we include batch normalization and $L_1$ regularization into the network training. 
    \item In \cref{sec:res_bo}, now with an optimized procedure for \gls{QAP}, we would like to understand the relationship between structured (neuron-wise) and unstructured (synapse-wise) pruning.  
    These two concepts are often overloaded but reduce computational complexity in different ways.  
    To do this, we compare the unstructured pruning procedure we introduced in~\cref{sec:res_perf} to removing whole neurons in the network.  
    Structured pruning, or optimizing the hyperparameter choice of neural network nodes, is performed using a Bayesian Optimization approach introduced in \cref{sec:bo}.  
    \item In \cref{sec:res_aiq}, we make preliminary explorations to understand the extent to which \gls{QAP} is removing important synapses which may prevent generalizability of the model.
    While there are a number of ways to test this; in our case, we test generalizability by randomizing a fraction of the class labels and checking if we are still able to prune the same amount of weights from the network as in the non-randomized case.  
\end{itemize}

\vspace{1cm}
\subsection{QAP performance}
\label{sec:QAPres}

\begin{figure*}[tbh!]
    \centering
     \includegraphics[width=0.80\textwidth]{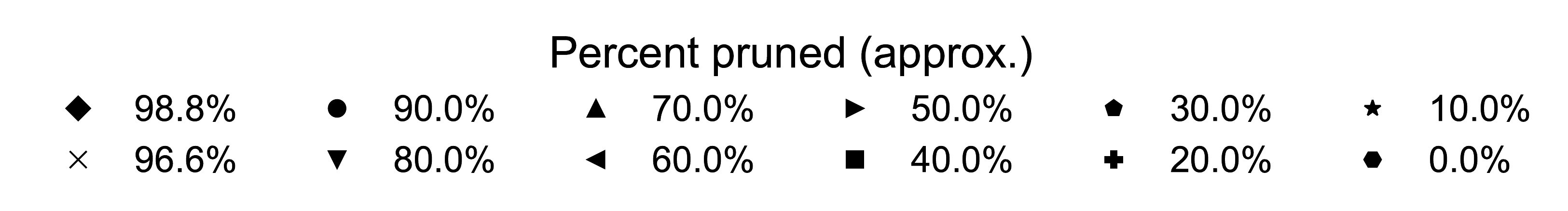}
    \includegraphics[width=0.48\textwidth]{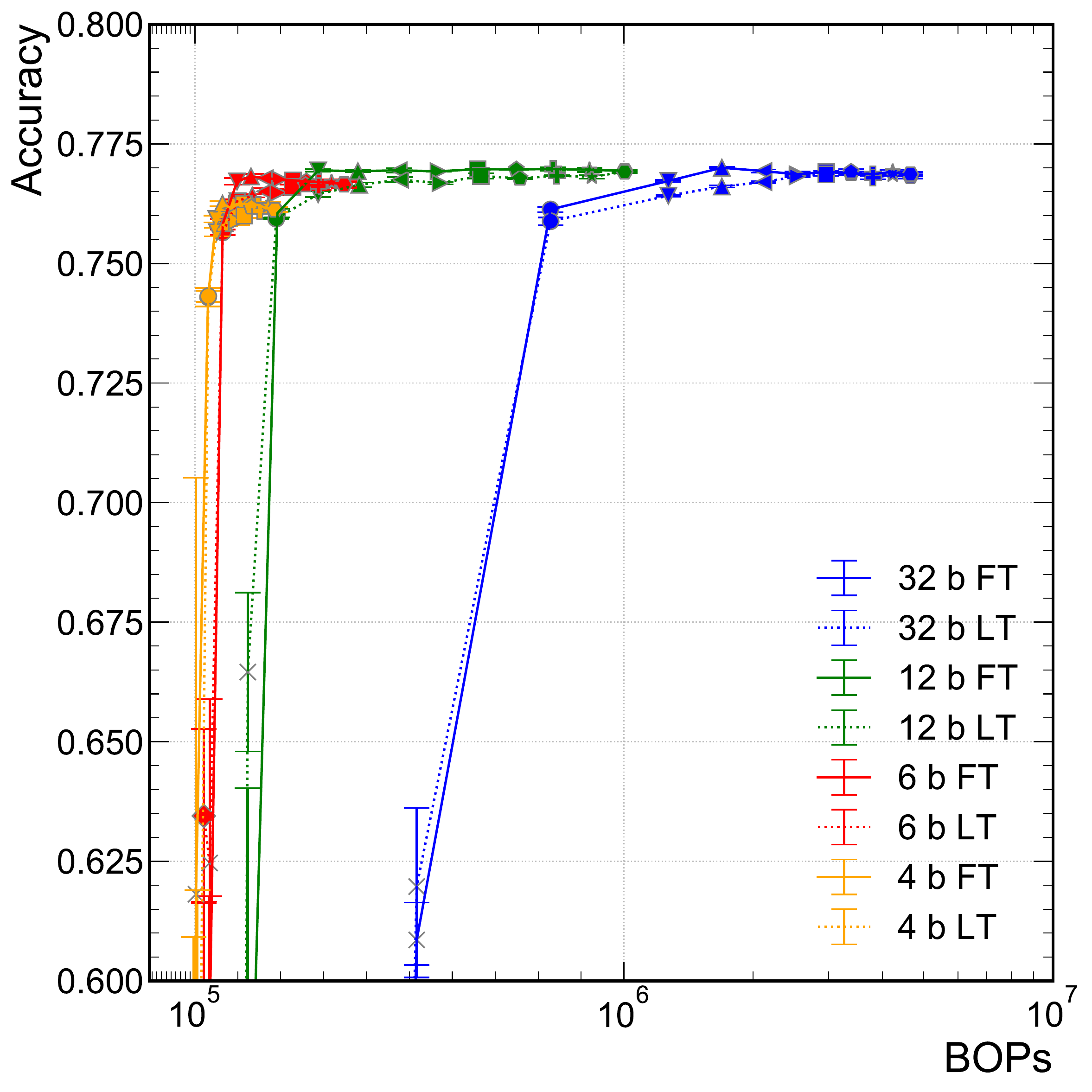}   
    \includegraphics[width=0.48\textwidth]{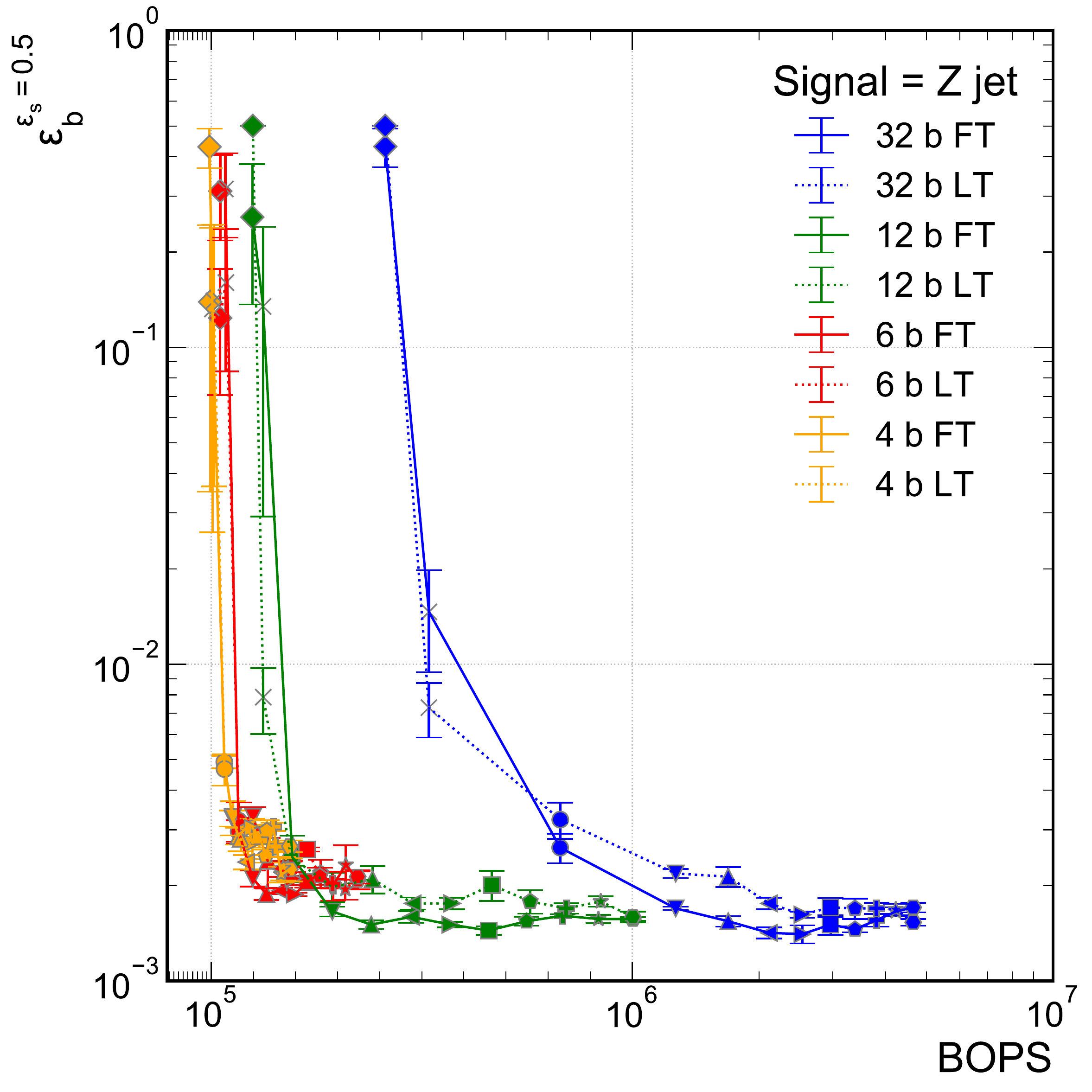}
    \caption{Model accuracy (A) and background efficiency (B) at 50\% signal efficiency versus \glspl{BOP} for different sparsities achieved via \gls{QAP}, for both \gls{FT} and \gls{LT} pruning techniques}
    \label{fig:precision}
\end{figure*}

The physics classifier performance is measured with the accuracy and $\eb$ metric for each signal class. 
We train a number of models at different precision: 32-bit floating-point precision and 12-, 6-, and 4-bit scaled-integer precision.  
For each precision explored, we then apply a pruning procedure.  
We explore both of the \gls{LT} and \gls{FT} pruning schemes described in \cref{sec:qap}.  
The result is illustrated in \cref{fig:precision} where each of the colored lines indicates a different model precision, the solid (dashed) lines correspond to \gls{FT} (\gls{LT}) pruning, and each of the points along the curves represents the percent of the original network weights that have been pruned.  
Each \gls{NN} includes a \gls{BN} layer after each of the hidden layers and has been trained including an $L_1$ regularization loss term. 
Further, each model's performance was verified via a $k$-fold cross-validation scheme, where $k=4$ in which training and validation datasets were shuffled over multiple training instances. Plotted performance is the mean value and error bars represent the standard error across the folds. 
All metrics were calculated on the same test dataset, which stayed static across each training instance.  
\label{sec:res_perf}
The first observation from \cref{fig:precision} is that we can achieve comparable performance to the 32-bit floating-point model with the 6-bit scaled-integer model.  
This is consistent with findings in a previous \qkeras-based study~\citep{Coelho:2020zfu} where, with uniform quantization, the performance was consistent down to 6-bit fixed-point quantization.  
When the precision is reduced to 4-bits, the performance begins to degrade.
Then, as we increasingly prune the models at all of the explored precisions, the performance is maintained until about 80\% of the weights are pruned.  
The observations are consistent whether we consider the accuracy (\cref{fig:precision} left) or $\eb$ (\cref{fig:precision} right) metric.
For the case of $\eb$, there is an increase of roughly 1.2--2$\times$ with respect to the 32-bit floating-point model; however, there are statistical fluctuations in the values because of the limited testing sample size and the small background efficiencies of $2 \times 10^{-3}$ that we probe.  
Instead, now if we compare the computational cost of our \gls{QAP} 6-bit model to the unpruned 32-bit model, we find a greater than $25\times$ reduction in computational cost (in terms of \glspl{BOP}) for the same classifier performance. 
For the jet substructure classification task, the quantization and pruning techniques are complementary and can be used in tandem at training time to develop an extremely efficient \gls{NN}.  
With respect to earlier work with \gls{FT} pruning at 32-bit floating-point precision and \gls{PTQ} presented in \cite{Duarte:2018ite}, we find a further greater than $3\times$ reduction in \glspl{BOP}.

In \cref{fig:precision}, we also find that there is no significant performance difference between using \gls{FT} and \gls{LT} pruning.
As we prune the networks to extreme sparsity, greater than 80\%, the performance begin to degrade drastically for this particular dataset and network architecture.  
While the plateau region is fairly stable, in the ultra-sparse region, there are significant variations in the performance metrics indicating that the trained networks are somewhat brittle.  
For this reason, we truncate the accuracy versus \glspl{BOP} graphs at 60\% accuracy.

\begin{figure*}[tbh!]
    \centering
    \includegraphics[width=0.48\textwidth]{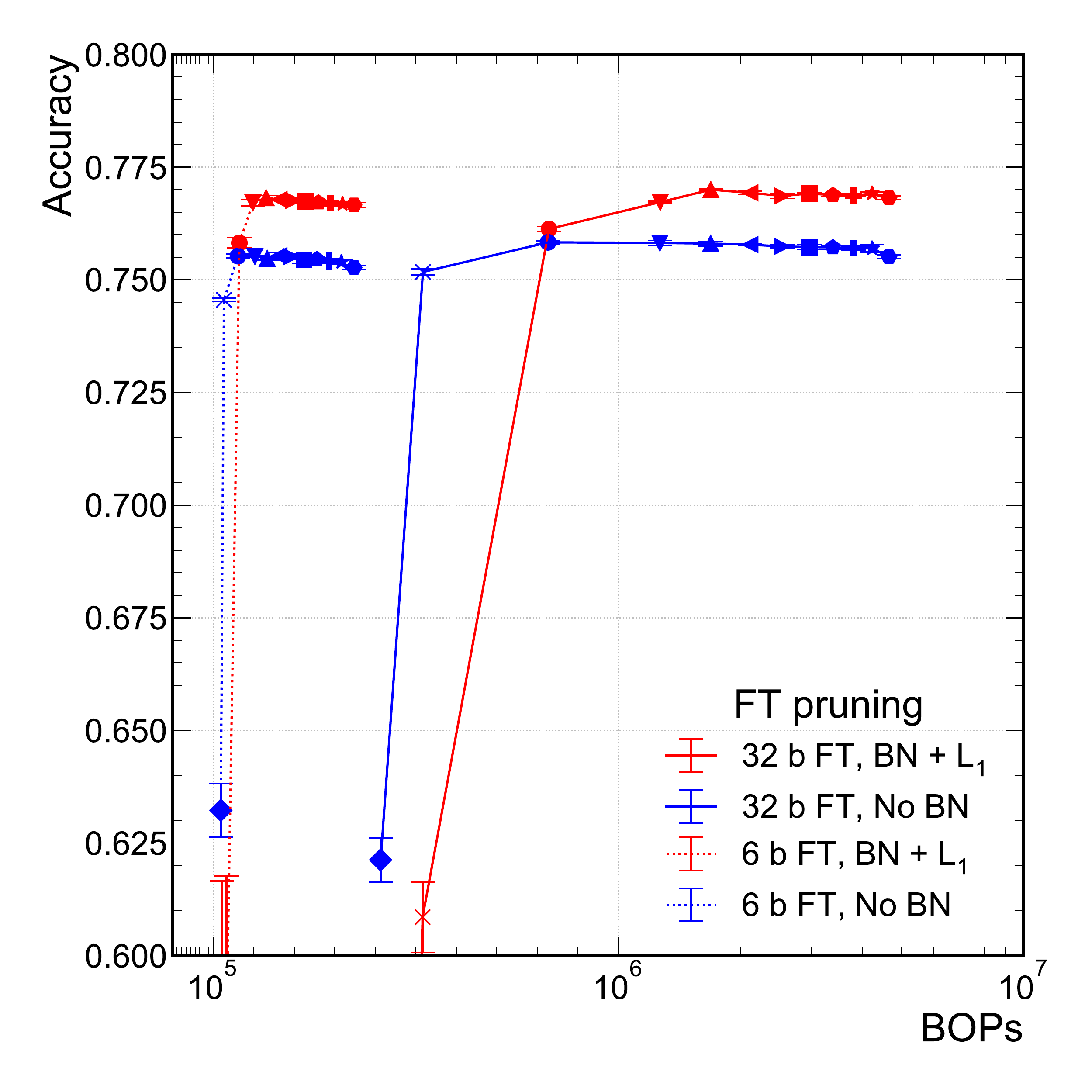}
    \includegraphics[width=0.48\textwidth]{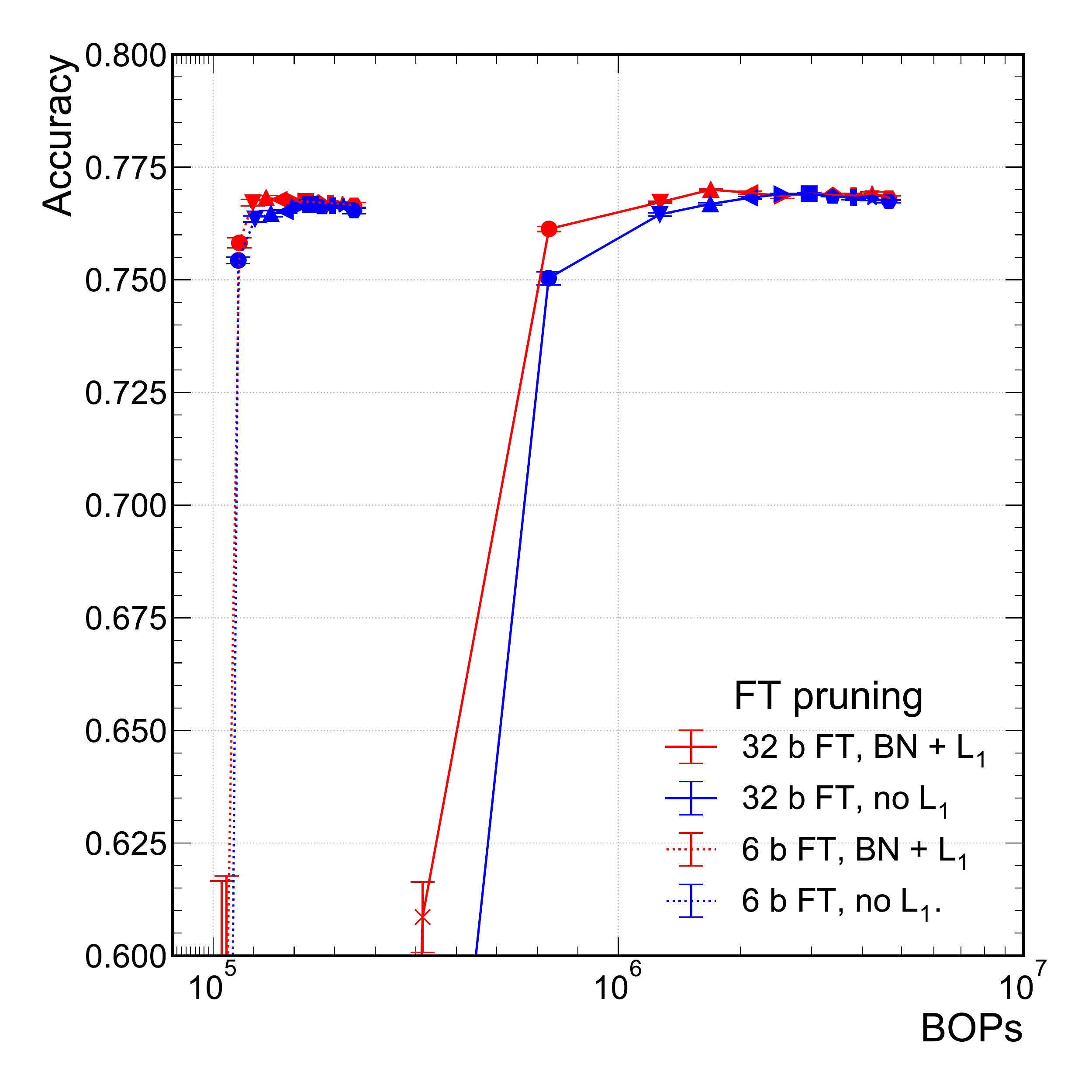}
    \caption{Comparison of the model accuracy when trained with \gls{BN} layers and $L_1$ regularization versus when trained without \gls{BN} layers (A) or $L_1$ regularization (B).}
    \label{fig:BNL1}
\end{figure*}

We also explore the performance of the model when removing either the \gls{BN} layers or the $L_1$ regularization term, which we term the \emph{no \gls{BN}} and \emph{no $L_1$} models, respectively.  
This is illustrated in \cref{fig:BNL1} for the 32-bit floating-point and 6-bit scaled-integer models.  
For easier visual comparisons, we omit the 4-bit and 12-bit models because the 6-bit model is the lowest precision model with comparable performance to the 32-bit model.  
In \cref{fig:BNL1} (A), we see that there is a modest performance degradation in the no \gls{BN} configuration for both lower and full precision models. 
In our application, we find that batch normalization does stabilize and improve the performance of our neural network and thus include it in our baseline model definition. 
In \cref{fig:BNL1} (B), we find that including or removing the $L_1$ regularization term in the loss function does not affect the performance significantly until extreme sparsity where the variations in performance can be large.  
However, as we will see in \cref{sec:res_aiq}, this does not mean that the entropic information content of the \glspl{NN} are similar.  

To highlight the performance of the \gls{QAP} procedure, we summarize our result compared to previous results for this jet substructure classification task with the same \gls{NN} architecture shown in \cref{fig:architectures}.
The results are summarized in \cref{tab:summary1}.  
In the nominal implementation, no quantization or pruning is performed. 
In \cite{Duarte:2018ite}, the 32-big floating-point model is \gls{FT} pruned and then quantized post-training.
This approach suffers from a loss of performance below 16 bits.  
Using \gls{QAT} and \qkeras~\citep{Coelho:2020zfu}, another significant improvement was demonstrated with a 6-bit fixed-point implementation.  
Finally, in this work with \gls{QAP} and \brevitas, we are able to prune the 6-bit network by another 80\%.
With respect to the nominal implementation we have reduced the \glspl{BOP} by a factor of 25, the original pruning + \gls{PTQ} approach a factor of 3.3, and the \gls{QAT} approach by a factor of 2.2.  

One further optimization step is to compare against a mixed-precision approach where different layers have different precisions~\citep{Coelho:2020zfu}.  
We leave the study of mixed-precision \gls{QAP} to future work and discuss it in \cref{sec:outlook}.\\
\newcommand\T{\rule{0pt}{2.6ex}}       
\newcommand\B{\rule[-1.2ex]{0pt}{0pt}} 
\begin{table*}[tbh!]
    \centering
    \caption{Performance evolution of the jet substructure classification task for this \gls{NN} architecture. 
    ``Nominal'' refers to an unpruned 32-bit implementation, ``pruning + \gls{PTQ}'' refers to a network with \gls{FT} pruning at 32-bit precision with \gls{PTQ} applied to reduce the precision to 16 bits, ``\gls{QAT}'' refers to a \qkeras implementation, and ``\gls{QAP}'' is this result. 
    The bolded value in each column indicates the best value of each metric. }
    \resizebox{\textwidth}{!}{
    \begin{tabular}{lllrrrrr}\hline
        Model \T\B& Precision & \gls{BN} or $L_1$ & Pruned [\%] & \glspl{BOP} & Accuracy [\%] & $\langle\eb\rangle$ [\%] & $\langle$AUC$\rangle$ [\%]\\
         \hline
         Nominal \T& 32-bit floating-point & $L_1$ + BN  & 0 & 4,652,832 & \textbf{76.977} & \textbf{0.00171} & \textbf{94.335} \\
         Pruning + \gls{PTQ} & 16-bit fixed-point & $L_1$ + BN & 70 & 631,791 & 75.01 & 0.00210 & 94.229\\
         \gls{QAT} & 6-bit fixed-point & $L_1$ + BN & 0 & 412,960 & 76.737 & 0.00208 & 94.206  \\
         \gls{QAP} \B& 6-bit scaled-integer & $L_1$ + BN & \textbf{80} & \textbf{189,672} & 76.602 & 0.00211 & 94.197 \\\hline
    \end{tabular}}
    \normalsize
    \label{tab:summary1}
\end{table*}


\subsection{Pruned versus unpruned quantized networks}
\label{sec:res_bo}

To compare against the efficacy of applying \gls{QAP}, we explore \gls{QAT} with no pruning.  
In an alternate training strategy, we attempt to optimize the \gls{NN} architecture of the unpruned \gls{QAT} models. 
This is done using the \gls{BO} technique presented in \cref{sec:opts}.  
The widths of the hidden layers are varied to find optimal classifier performance. 
We compare the performance of this class of possible models using \gls{BO} against our \gls{QAP} procedure, including \gls{BN} and $L_1$ regularization, presented in the previous section.
It is important to note, as we will see, that \gls{QAP} and \gls{BO} are conceptually different procedures and interesting to compare. 
The \gls{QAP} procedure starts with a particular accuracy-optimized model and attempts to ``streamline'' or compress it to its most optimal bit-level implementation.  
This is the reason that the accuracy drops precipitously when that particular model can no longer be streamlined.  
Alternatively, the family of \gls{BO} models explores the Pareto optimal space between \glspl{BOP} and accuracy.  
In future work, we would like to further explore the interplay between \gls{QAP} and \gls{BO}.  

\begin{figure*}[tbh!]
    \centering
    \includegraphics[width=0.45\textwidth]{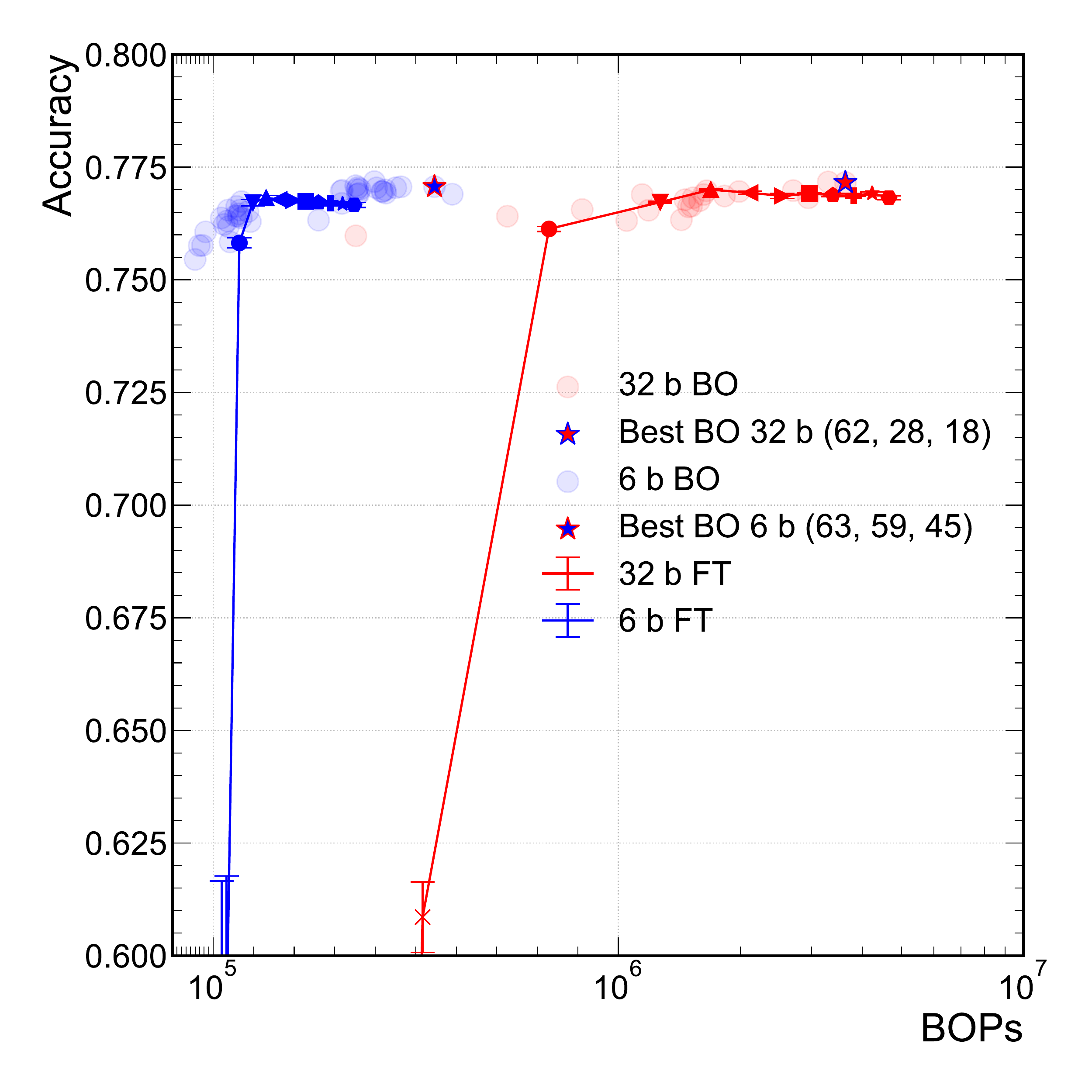}  
    \includegraphics[width=0.45\textwidth]{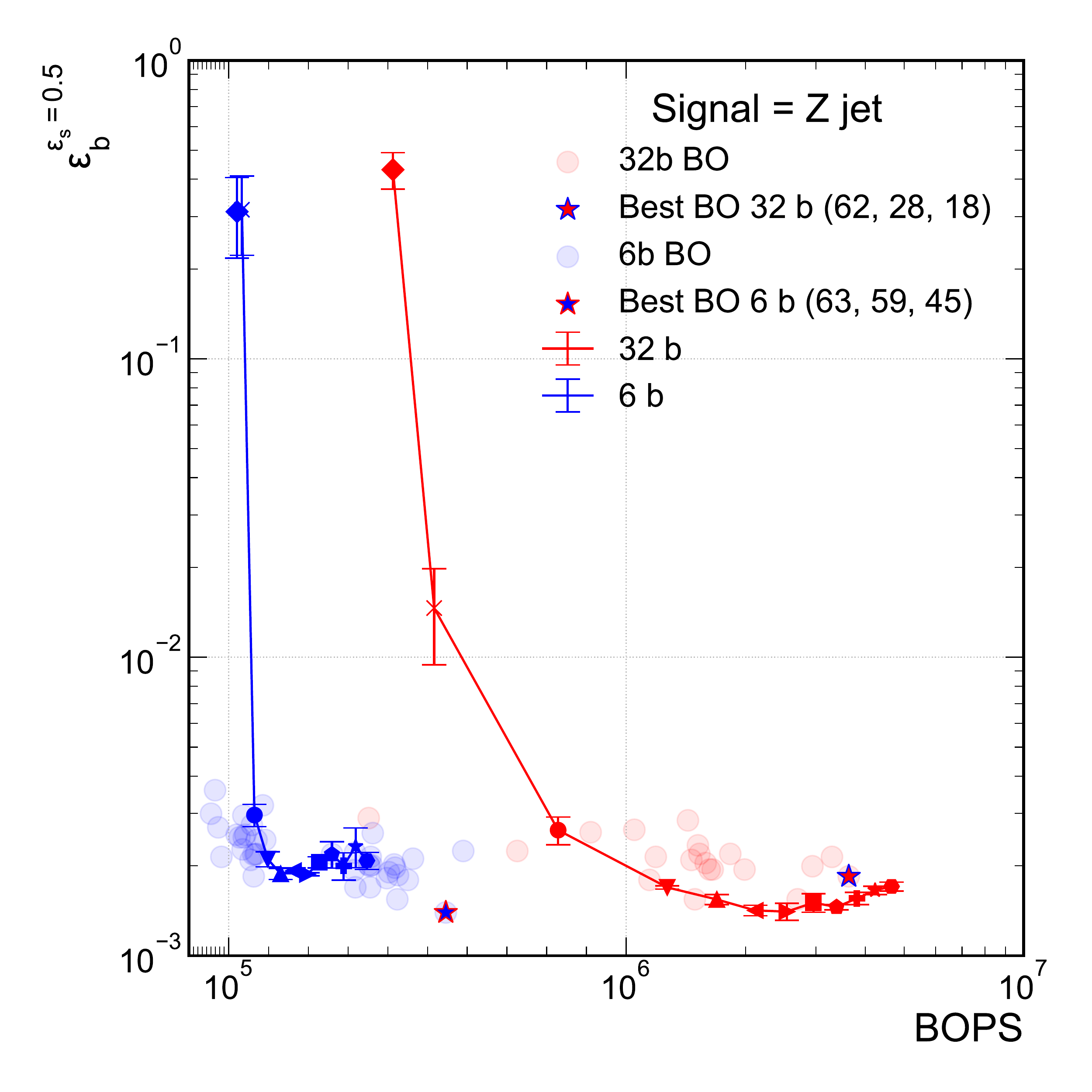}
    \caption{Comparison of \gls{FT} pruned model's and \gls{BO} model's accuracy (A) and background efficiency (B) at 50\% signal efficiency. 
    Each hyperparameter configuration that was explored during the \gls{BO} procedure is marked as a transparent dot, with the resulting ``best'' model, which the lowest BCE Loss as calculated on the ``test'' set, is marked by the outlined star.}
    \label{fig:BO}
\end{figure*}

\cref{fig:BO} presents both the accuracy versus \glspl{BOP} curves for the  \gls{QAP} models and the unpruned \gls{QAT} models found using \gls{BO}. 
For ease of comparison, we display only the 32-bit and 6-bit models.  
The solid curves correspond to the \gls{QAP} models while the individual points represent the various trained unpruned models explored during the \gls{BO} procedure.
The unpruned model with the highest classification performance found using the \gls{BO} procedure is denoted by the star.  
While the starred models are the most performant, there is a class of \gls{BO} models that tracks along the \gls{QAP} curves fairly well.
There is a stark difference in how \gls{QAP} and \gls{BO} models behave as the accuracy degrades below the so-called ``plateau'' region where the accuracy is fairly constant and optimal.  
When the sub-network of the \gls{QAP} model can no longer approximate the optimally performing model, its performance falls off dramatically and the accuracy drops quickly.   
Because \gls{BO} explores the full space including Pareto optimal models in \glspl{BOP} versus accuracy, they exhibit a more gentle decline in performance at small values of \glspl{BOP}.  
It is interesting to note that the classification performance of the \gls{BO} models begins to degrade where the \gls{QAP} procedure also falls off in performance; for example, just above $10^5/$\glspl{BOP} in \cref{fig:BO}A for the 6-bit models.  
We anticipate future work to explore combining \gls{BO} and \gls{QAP} procedures to see if any accuracy optimal model can be found at smaller \glspl{BOP} values.  


\vspace{1cm}
\subsection{Entropy and generalization}
\label{sec:res_aiq}
\Gls{QAP} models exhibit large gains in computational efficiency over (pruned and unpruned) 32-bit floating-point models, as well as significant gains over unpruned \gls{QAT} models for our jet substructure classification task.  
In certain training configurations, we have found similar performance but would like to explore if the information in the neural network is \emph{represented} similarly.  
As a metric for the information content of the \gls{NN}, we use the \emph{neural efficiency} metric defined in \cref{eq:neff}, the Shannon entropy normalized to the number of neurons in a layer then averaged over all the layers of the \gls{NN}.  

By itself, the neural efficiency is an interesting quantity to measure.
However, we specifically explore the hypothesis, described in \cref{sec:metrics}, that the neural efficiency is related to a measure of generalizability.  
In this study, we use the classification performance under different rates of class randomization during training as a probe of the generalizability of a model.  
We randomize the class labels among the five possible classes for 0\%, 50\%, 75\%, and 90\% of the training dataset. 
To randomize the training data, we iterate over a given percent of the normal dataset, setting the real class of each input to 0, choosing a new class at random out of the 5 possible, then setting that new class to 1. 
The data is then shuffled and split as normal.

\begin{figure*}[tbh!] 
    \centering
    \includegraphics[width=0.7\textwidth]{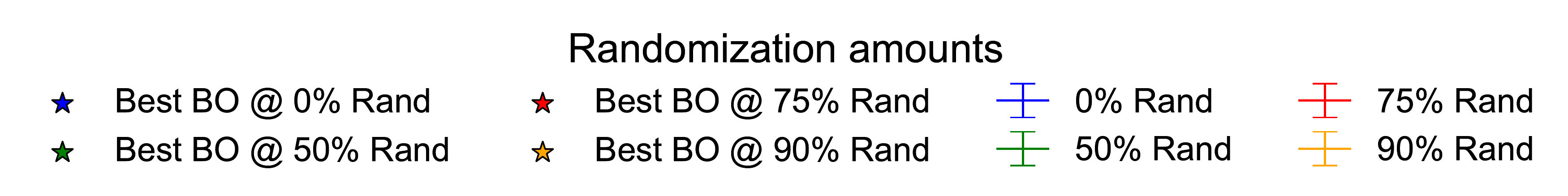}\\
    \includegraphics[width=0.32\textwidth]{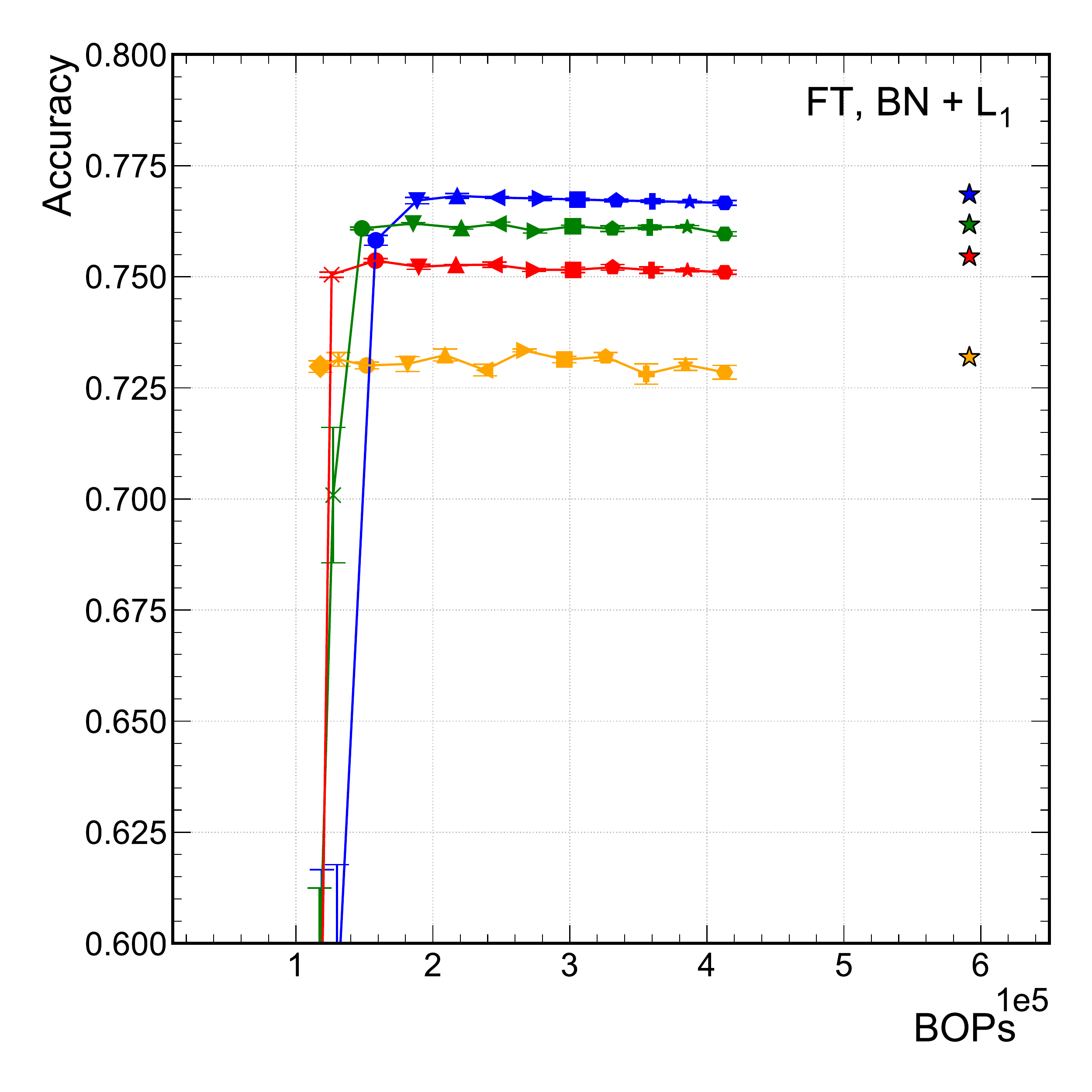}
    \includegraphics[width=0.32\textwidth]{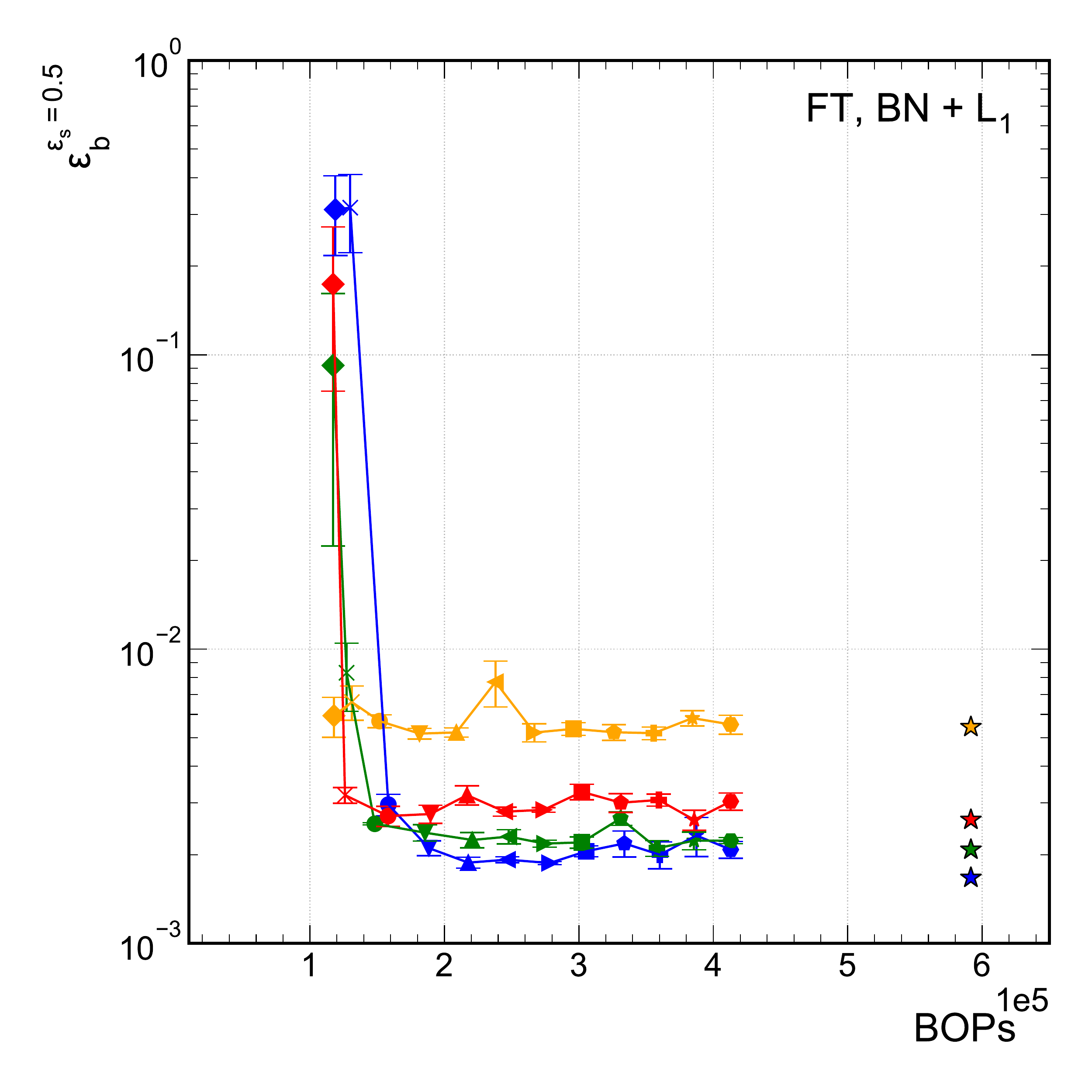}    \includegraphics[width=0.32\textwidth]{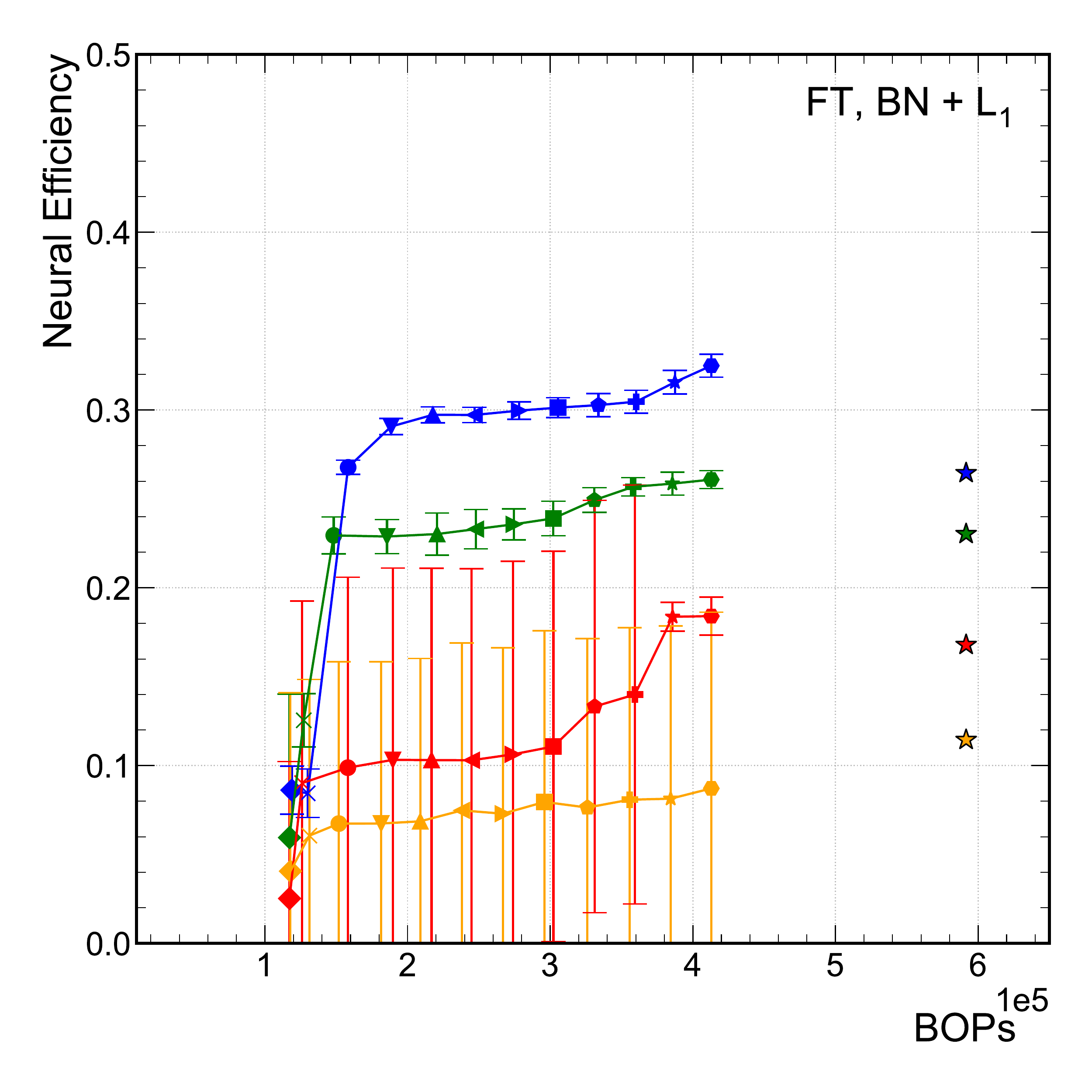}\\
    \includegraphics[width=0.32\textwidth]{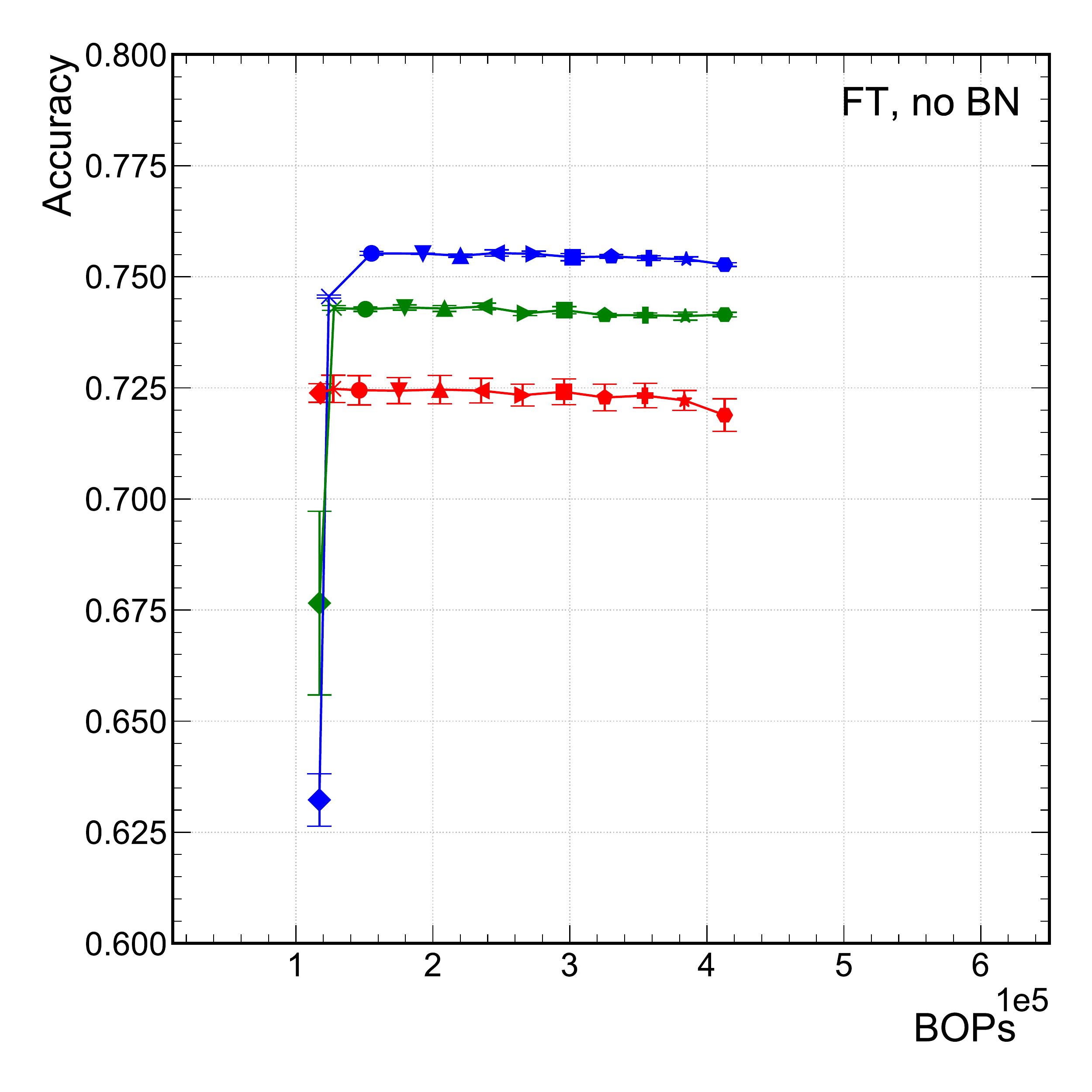}
    \includegraphics[width=0.32\textwidth]{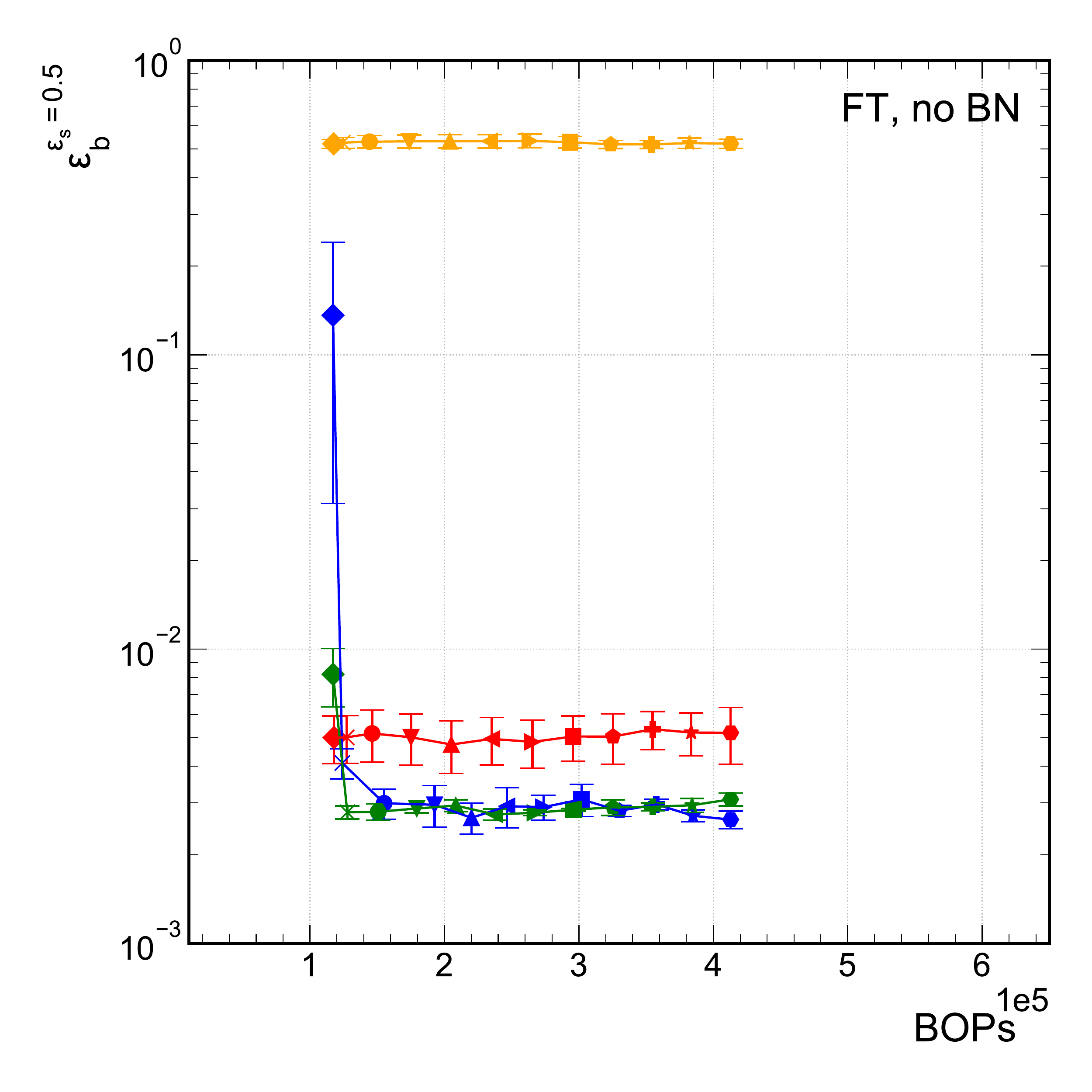}    \includegraphics[width=0.32\textwidth]{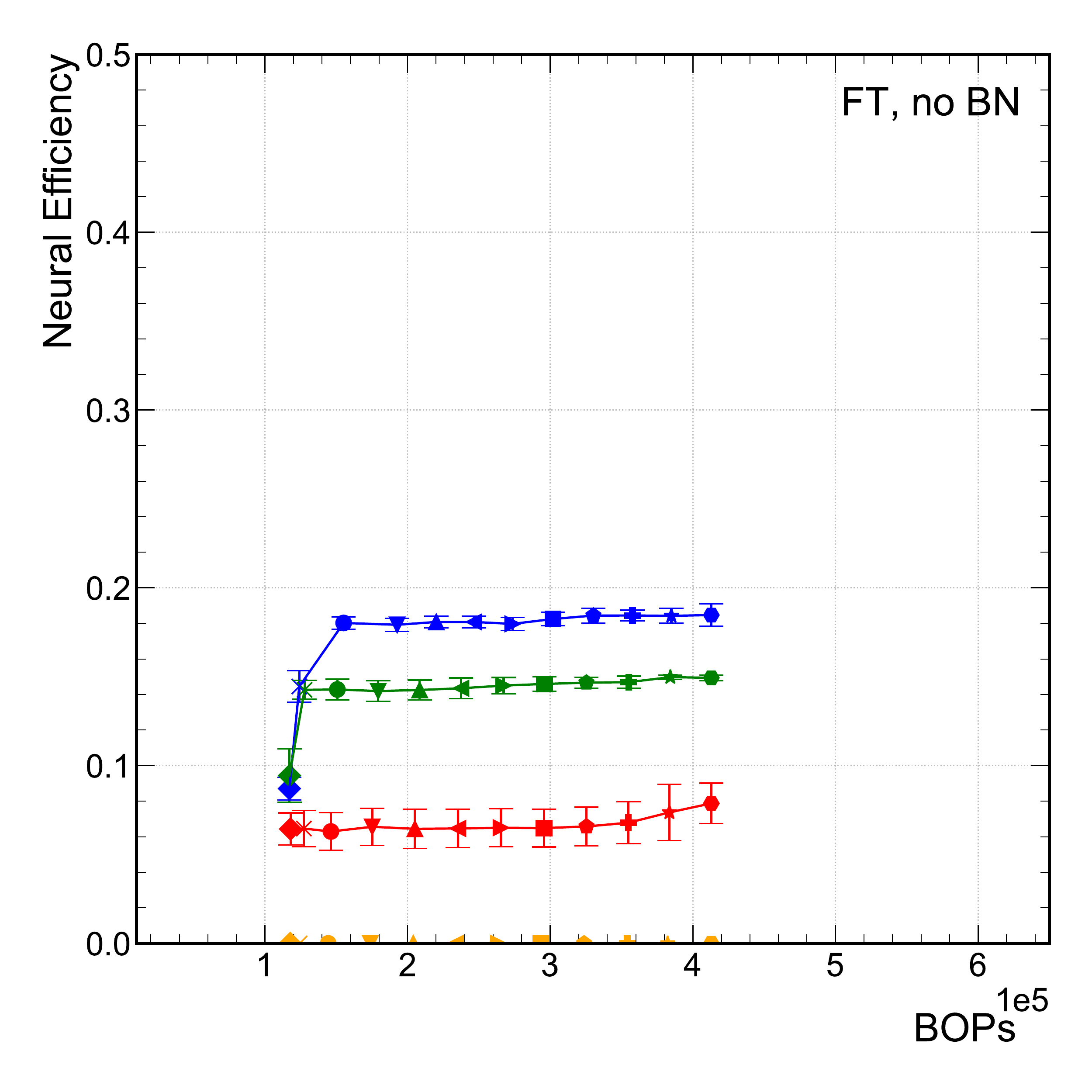}\\
    \includegraphics[width=0.32\textwidth]{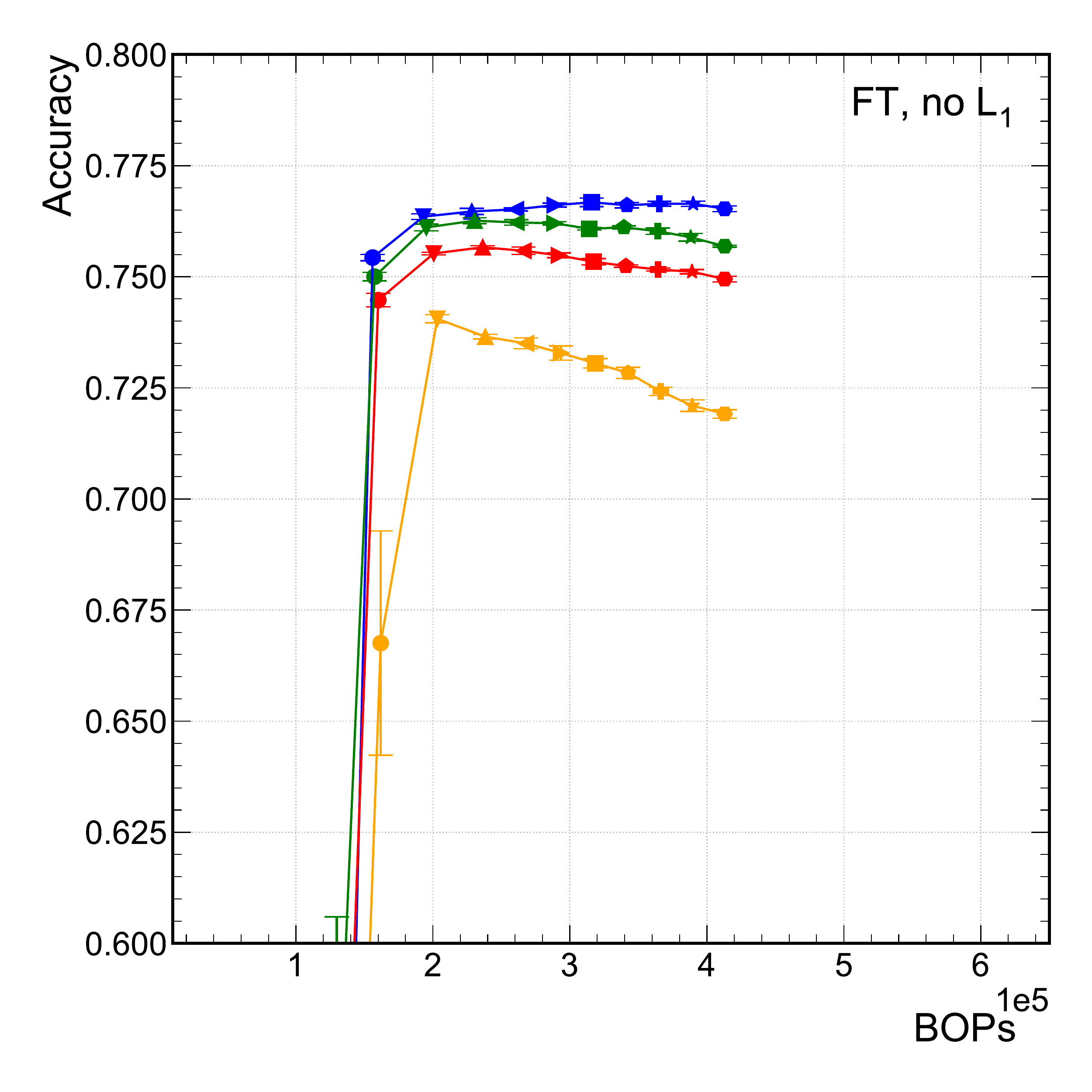}    
    \includegraphics[width=0.32\textwidth]{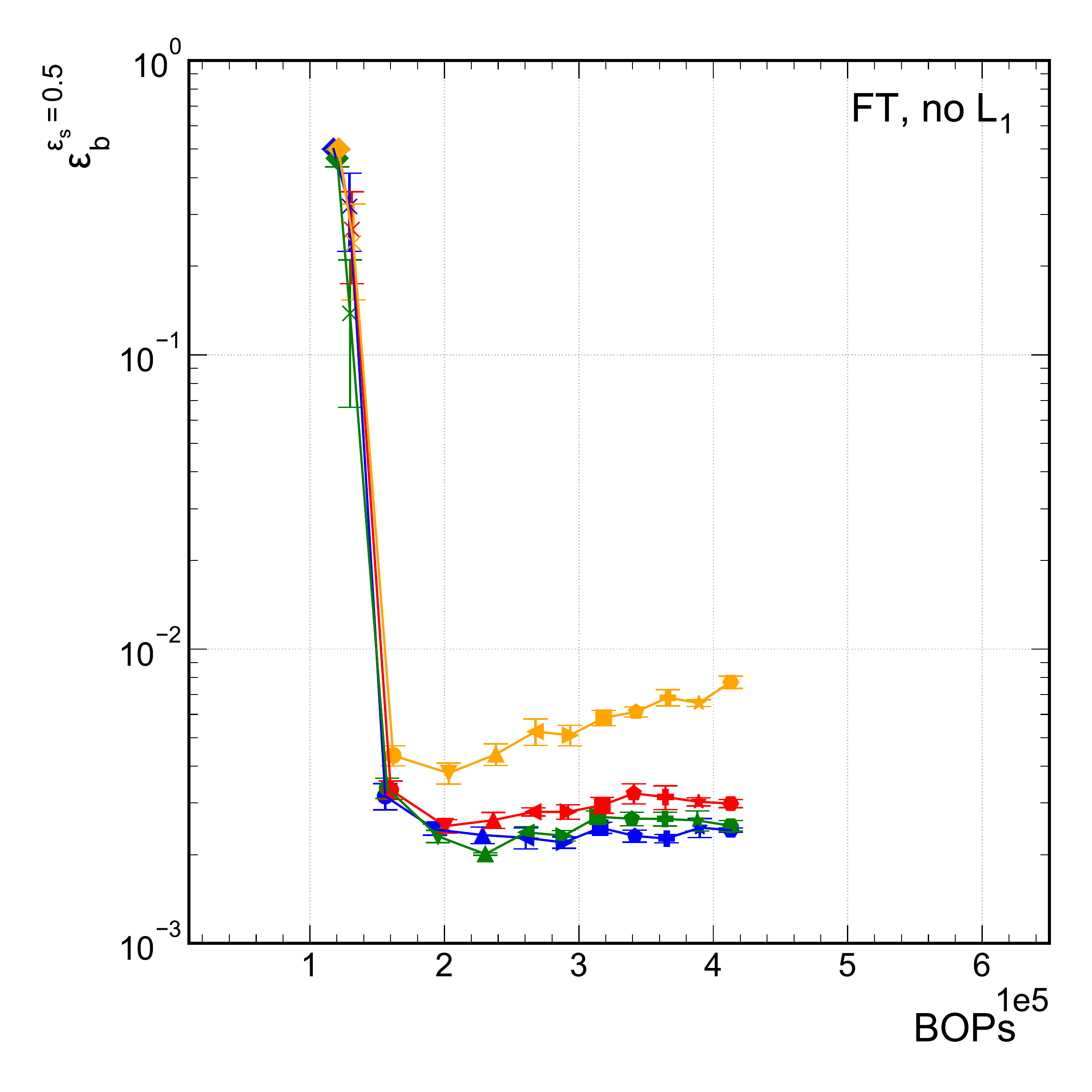}    \includegraphics[width=0.32\textwidth]{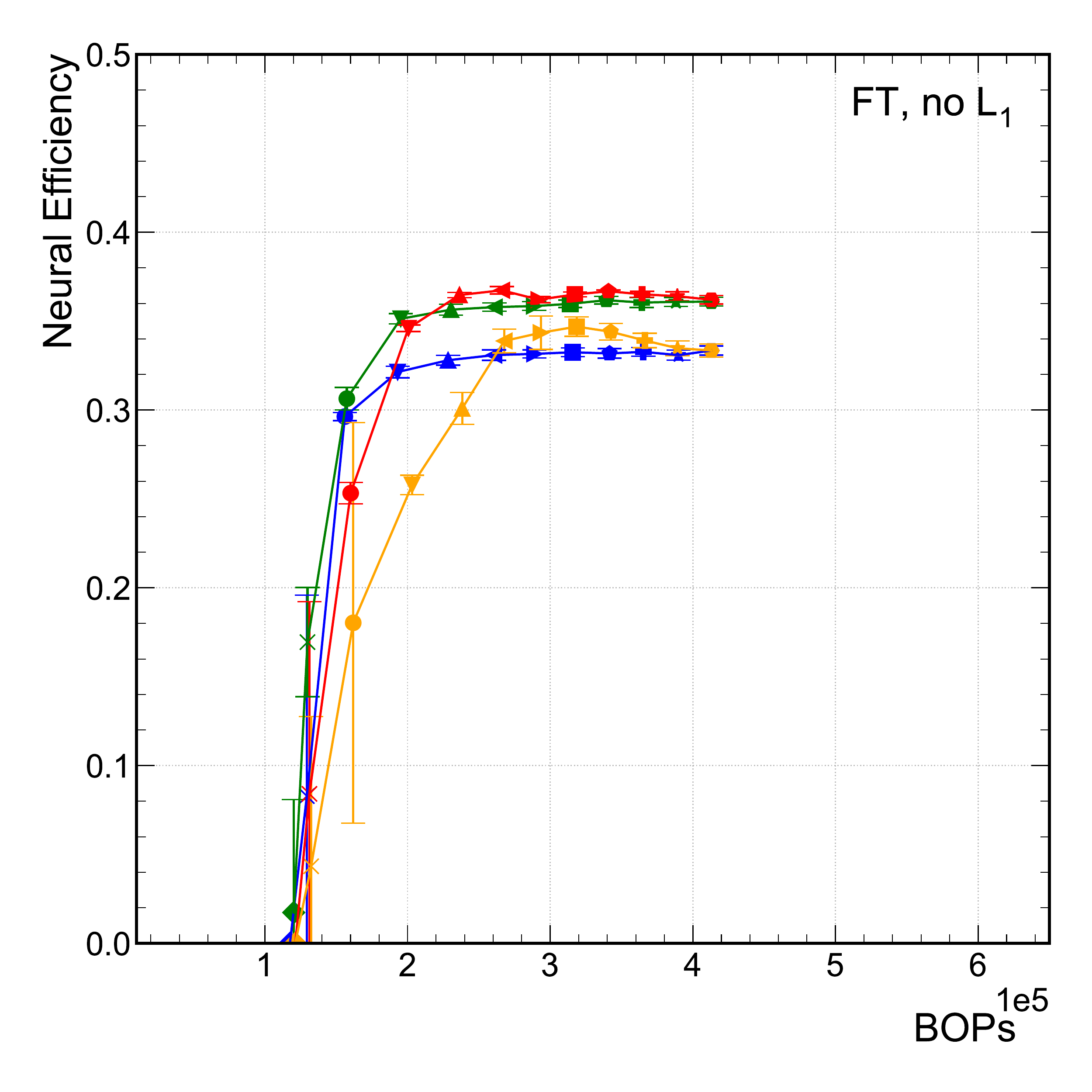}\\
    \caption{ Comparison of accuracy, $\eb$, and neural efficiency at 50\% signal efficiency for a 6-bit QAP model as BN layers and/or $L_1$ regularization is present in the model. 
    $L_1$ + \gls{BN} (upper), no \gls{BN} (middle), and no $L_1$ (lower)}
    \label{fig:accneff}
\end{figure*}

To compare with the results in \cref{sec:QAPres}, we study models that are trained using \gls{QAP} with 6-bit precision and are pruned using the fine-tuning pruning procedure.  
The results are presented in \cref{fig:accneff} where the left column shows the classifier accuracy versus \glspl{BOP}.  
The center column shows the $\eb$ metric. 
The right column displays the neural efficiency versus \glspl{BOP}.
The three rows explore three different scenarios: with both \gls{BN} and $L_1$ regularization (upper), no \gls{BN} (middle), and no $L_1$ (lower).
The various curves presented in each graph correspond to different class label randomization fractions of the training sample.

Among these training procedures, the $L_1$ + \gls{BN} model accuracy (upper left) is the highest and most consistent across the entire pruning procedure.  
Even with 90\% class randomization, the accuracy is still greater than 72.5\% and $\eb < 10^{-2}$. 
Alternatively, the no \gls{BN} model accuracy is consistently worse than the $L_1$ + \gls{BN} models for all values of randomization. 
Interestingly, the no \gls{BN} model accuracy with 90\% randomization drops precipitously out of the range of the graphs indicating that \gls{BN} is even more important to performance 
when class randomization is introduced.  
Meanwhile, the no $L_1$ model exhibits an interesting behavior with lower accuracy at larger values of \glspl{BOP}.  
As the no $L_1$ model is pruned, the accuracy improves until we arrive at extreme sparsity and the model performance degrades as usual.  
Our interpretation is that the generalization power of the unregularized model is worse than the $L_1$ regularized models.  
However, as we implement the \gls{QAP} procedure, the pruning effectively regularizes the model building robustness to the class randomization and recovering some of the lost accuracy.

The corresponding neural efficiency plots are shown in the right column of \cref{fig:accneff}.  
As a general observation, we find that the neural efficiency follows the same trend versus \glspl{BOP} as the accuracy, i.e. that within a given training configuration, the neural efficiency is stable up to a given sparsity.
Thus, up to this point, pruning does not affect the information content.  
This is particularly true in the case of the no \gls{BN} model, while with \gls{BN} there is more freedom, and thus modest variation in neural efficiency during the pruning procedure.

If we first only consider the 0\% randomized models for the right column, we can  see that the neural efficiency drops from about 0.3 to about 0.2 with the no \gls{BN} configuration.
As the neural efficiency is a measure of how balanced the neurons are activated (i.e. how efficiently the full state space is used), we hypothesize that \gls{BN} more evenly distributes the activation among neurons.
For the models that include $L_1$ regularization (upper and middle), the neural efficiency drops along with the accuracy as the randomization is increased. 
This effect is not nearly as strong in the no $L_1$ case in the lower row. 
We note that the performance of the 90\% randomized no \gls{BN} model is catastrophically degraded and the neural efficiency drops to zero, which we interpret to indicate that \gls{BN} is an important factor in the robustness and generalizability of the model.  

The no $L_1$ models (lower) are particularly notable because the neural efficiency does not decrease much as we the class randomization fraction is increased, in contrast with the upper and middle rows of \cref{fig:accneff}.  
This however, does not translate into a more robust performance. 
In fact, at 90\% class randomization and 80\% pruned, the $L_1$ + \gls{BN} and no $L_1$ models are drastically different in neural efficiency while being fairly similar in classifier accuracy.

Finally, the accuracy and neural efficiency of the highest accuracy models from the \gls{BO} procedure in \cref{sec:res_bo} are represented as stars in the top row of \cref{fig:accneff}. 
They have slightly lower neural efficiencies because the width of each hidden layer is bigger than in the \gls{QAP} models while the entropy remains relatively similar to those same models.  
The \gls{BO} models, as seen in the upper left graph of \cref{fig:accneff}, are no better at generalizing under increasing class randomization fractions than the \gls{QAP} models.

%% file: outlook.tex
\label{sec:outlook}
\glsresetall 

In this study, we explored efficient \gls{NN} implementations by coupling pruning and quantization at training time.  
Our benchmark task is ultra low latency, resource-constrained jet classification in the real-time online filtering system, implemented on \glspl{FPGA}, at the \gls{LHC}.  
This classification task takes as inputs high-level expert features in a fully-connected \gls{NN} architecture.  

Our procedure, called \gls{QAP}, is a combination of \gls{QAT} followed by iterative unstructured pruning.  
This sequence is motivated by the fact that quantization has a larger impact on a model's computational complexity than pruning as measured by \glspl{BOP}.  
We studied two types of pruning: \gls{FT} and \gls{LT} approaches.  
Furthermore, we study the effect of \gls{BN} layers and $L_1$ regularization on network performance.  
Under this procedure, considering networks with uniformly quantized weights, we found that with nearly no loss in classifier accuracy and 1.2--2$\times$ increase in \eb, the number of \glspl{BOP} can be reduced by a factor of 25, 3.3, and 2.2 with respect to the nominal 32-bit floating-point implementation, pruning with \gls{PTQ}, and \gls{QAT}, respectively.  
This demonstrates that, for our task, pruning and \gls{QAT} are complementary and can be used in concert.  

Beyond computational performance gains, we sought to understand two related issues to the \gls{QAP} procedure.  
First, we compare \gls{QAP} to \gls{QAT} with a \gls{BO} procedure that optimizes the layer widths in the network.  
We found that the \gls{BO} procedure did not find a network configuration that maintains performance accuracy with fewer \glspl{BOP} and that both procedures find similarly efficiently sized networks as measured in \glspl{BOP} and high accuracy.  

Second, we studied the information content, robustness, and generalizability of the trained \gls{QAP} models in various training configurations and in the presence of randomized class labels.
We compute both the networks' accuracies and their entropic information content, measured by the neural efficiency metric~\citep{aiq}.
We found that both $L_1$ regularization and \gls{BN} are required to provide the most robust \glspl{NN} to class randomization.  
Interestingly, while removing $L_1$ regularization did not significantly degrade performance under class randomization, the neural efficiencies of the \glspl{NN} were vastly different---varying by up to a factor of 3.  
This illustrates, that while \glspl{NN} may arrive at a similar performance accuracy, the information content in the networks can be very different.  \\

\subsection{Outlook}

As one of the first explorations of pruning coupled with quantization, our initial study of \gls{QAP} lends itself to a number of follow-up studies.  

\begin{itemize}
    \item Our benchmark task uses high-level features, but it is interesting to explore other canonical datasets, especially those with raw, low-level features. 
    This may yield different results, especially in the study of generalizability.
    \item Combining our approach with other optimization methods such as Hessian-based quantization~\citep{hawq,hawqv2} and pruning could produce networks with very different \glspl{NN} in information content or more optimal solutions, particularly as the networks become very sparse.
    \item An important next step is evaluating the actual hardware resource usage and latency of the \gls{QAP} \glspl{NN} by using \gls{FPGA} co-design frameworks like \hlsfml~\citep{Duarte:2018ite} and FINN~\citep{finn,blott2018finnr}.  
    \item It would be interesting to explore the differences between seemingly similar \glspl{NN} beyond neural efficiency; for example, using metrics like singular vector canonical correlation analysis (SVCCA)~\citep{raghu2017svcca} which directly compare two \glspl{NN}
    \item We would like to explore further optimal solutions by combining \gls{BO} and \gls{QAP} procedures.  
    Beyond that, there is potential for more efficient solutions using mixed-precision \gls{QAT}, which could be done through a more general \gls{BO} procedure that explores the full space of layer-by-layer pruning fractions, quantization, and sizes.
\end{itemize}


\gls{QAP} is a promising technique to build efficient \gls{NN} implementations and would benefit from further study on additional benchmark tasks.  
Future investigation of \gls{QAP}, variations on the procedure, and combination with complementary methods may lead to even greater \gls{NN} efficiency gains and may provide insights into what the \gls{NN} is learning.  
